\documentclass{article}
\usepackage{amsmath,amssymb}
\usepackage{natbib,geometry}                % See geometry.pdf to learn the layout options. There are lots.
\usepackage{graphicx}
\usepackage{enumitem}
\usepackage{booktabs}
\newcommand{\scp}[2]{{\big\langle {#1}\, , \, {#2}\big\rangle}}

\newcommand{\mR}{{\mathbb R}}
\newcommand{\id}{{\mathrm{id}}}
\newcommand{\tht}{{\theta}}
\newcommand{\Ga}{{\Gamma}}
\newcommand{\ka}{\kappa}
\newcommand{\Diff}{{\mathrm{Diff}}}
\newcommand{\prt}{\partial}
\newcommand{\bsu}{{\boldsymbol u}}
\newcommand{\bsa}{{\boldsymbol a}}
\newcommand{\bsp}{{\boldsymbol p}}
\newcommand{\bsz}{{\boldsymbol z}}
\renewcommand{\phi}{\varphi}
\newcommand{\Sig}{\Sigma}
\newcommand{\Id}{\mathrm{Id}}
\newcommand{\hlineadd}{\midrule}

\graphicspath{{Figures/}}
\bibpunct{(}{)}{;}{a}{,}{,}

\begin{document}

\title{Diffeomorphic Learning}
\author{Laurent Younes\\
Department of Applied Mathematics and Statistics and Center for Imaging Science\\  Johns Hopkins University\\
%3400 N.Charles st. \\
%Baltimore, MD 21209\\
laurent.younes@jhu.edu}
\date{\today}
%\editor{}

\maketitle

\begin{abstract}
We introduce in this paper a learning paradigm in which the training data is transformed by a diffeomorphic transformation before prediction. The learning algorithm minimizes a cost function evaluating the prediction error on the training set penalized by the distance between the diffeomorphism and the identity. The approach borrows ideas from shape analysis where diffeomorphisms are estimated for shape and image alignment, and brings them in a previously unexplored setting, estimating, in particular diffeomorphisms in much larger dimensions. After introducing the concept and describing a learning algorithm, we present diverse applications, mostly with synthetic examples, demonstrating the potential of the approach, as well as some insight on how it can be improved. 

{ \bf keywords:}
Diffeomorphisms, Reproducing kernel Hilbert Spaces, Classification 

\end{abstract}

\section{Introduction}
We consider, in this paper, a family of classifiers that take the form  $x\mapsto F(\psi(x))$, $x\in \mR^d$, where $\psi$ is a diffeomorphism of $\mathbb R^d$ and $F$ is real-valued or categorical, belonging to a class of simple (e.g., linear) predictors.  We will describe a training algorithm that, when presented with a finite number of training samples, displaces all points together through non-linear trajectories, in order to bring them to a position for which the classes are linearly separable. Because these trajectories are built with a guarantee to avoid collisions between points, they  can be interpolated to provide a fluid motion of the whole space ($\mathbb R^d$) resulting in a diffeomorphic transformation. One can then use this transformation to assign a class to any new observation. Because one may expect that the simplicity of the transformation will be directly related to the ability of the classifier to generalize, the point trajectories and their interpolation are estimated so that this resulting global transformation is penalized from erring too far away from the identity, this penalty being assessed using a geodesic distance in the group of diffeomorphisms, based on methods previously developed for shape analysis.

\bigskip

The last decade's achievements in deep learning have indeed demonstrated that optimizing nonlinear transformations of the data in very high dimensions and using massive parametrization could lead to highly performing predictions without being necessarily struck by the curse of dimensionality. 
%Recent attempts at elucidating the reasons for such successes suggest that learning algorithms for neural nets, through their initialization or optimization strategies, create an implicit regularization that drive the procedures towards good solutions, even though understanding the exact nature of this regularization is still open. 
The approach that we describe here also explores a ``very large'' space in terms, at least, of the number of parameters required to describe the transformations, and uses a training algorithm that implements, like neural networks, dynamic programming.  It however frames the estimated transformations within a well specified space of diffeomorphisms, whose nonlinear structure is well adapted to the successive  compositions of functions that drive deep learning methods. While it shares some of the characteristics of  deep methods, our formulation relies on a regularization term, which makes it closer in spirit with many of the methods used for non-parametric prediction.
\bigskip

We now sketch our model for classification, and for this purpose introduce some notation. We let $c$ denote the number of classes for the  problem at hand. As a consequence, the function $F$ can either take values in $C = \{0, \ldots, c-1\}$, or in the space of probability distributions on $C$ (which will be our choice since we will use logistic regression). We furthermore assume that $F$ is parametrized,  by a parameter $\theta\in \mR^q$, and we will write $F(x; \theta)$ instead of $F(x)$ when needed.

Assume that a training set is given, in the form $\mathcal T_0 = (x_1, y_1, \ldots, x_N, y_N)$ with $x_k\in \mR^d$ and $y_k\in C$, for $k=1, \ldots, N$. We will assume that this set is not redundant, i.e., that $x_i\neq x_j$ whenever $i\neq j$. (Our construction can be extended to redundant training sets by allowing the training class variables $y_k$ to be multi-valued. This would result in rather cumbersome notation that is better avoided.) If $\psi$ is a diffeomorphism of $\mR^d$, we will let $\psi\cdot\mathcal T_0 = (\psi(x_1), y_1, \ldots, \psi(x_N), y_N)$.
The learning procedure will consist in minimizing the objective function
\begin{equation}
\label{eq:cost}
G(\psi, \theta) = D(\id, \psi)^2 + \lambda g(\tht) + \frac1{\sigma^2} \Gamma(F(\cdot, \theta), \psi\cdot\mathcal T_0)
\end{equation}
with respect to $\psi$ and $\theta$. Here $D$ is a Riemannian distance in a group of diffeomorphisms of $\mR^d$, that will be described in the next section, $\Gamma$ is a ``standard'' loss function, $g$ a regularization term on the parameters of the final classifier and $\lambda$, $\sigma$ are positive numbers. One can take for example
\begin{equation}
\label{eq:Log.reg}
\Gamma(F(\cdot, \theta), \psi\cdot \mathcal T_0) = - \sum_{k=1}^N \log F(\psi(x_k);\theta)(y_k),
\end{equation}
where
\[
F(z, \theta)(y) = \frac{e^{\theta(y)^Tz}}{\sum_{y'\in C} e^{\theta(y')^Tz}}
\]
and where the parameter is $(\tht_1, \ldots, \tht_{c-1}) \in (\mR^d)^{c-1}$ with $\theta_0 = 0$, letting (as done in our experiments) $g(\theta) = |\tht|^2$.

The proposed construction shares some of its features with  feed-forward networks \citep{goodfellow2016deep}, for which, for example, using the loss function above on the final layer is standard, and can also be seen as  a kernel method,  sharing this property with classifiers such as support vector machines (SVMs) or other classification methods that use the ``kernel trick'' \citep{vapnik2013nature,ss02}. However, the algorithm  is directly inspired from diffeomorphic shape analysis, and can be seen as a variant of the Large Deformation Diffeomorphic Metric Mapping (LDDMM) algorithm, which has been introduced for image and shape registration \citep{jm00,bmty05,younes2010shapes}. Up to our knowledge, this theory has not been applied so far to classification problems, although some similar ideas using a linearized model have been suggested in \citep{trouve2001metric} for the inference of metrics in shape spaces, and a similar approach has been discussed in \cite{walder2009diffeomorphic} in the context of dimensionality reduction.

While the underlying theory in shape analysis, which formulates the estimation of transformations as optimal control problems, is by now well established, the present paper introduces several original contributions. Translating shape analysis methods designed for small dimensional problems to larger scale problems indeed introduces new challenges, and suggests new strategies on the design of the flows that control the diffeomorphic transformations, on the choice of parameters driving the model and on optimization schemes. Importantly, the paper  provides experimental evidence  that diffeomorphic methods can be competitive in machine learning contexts. Indeed, diffeomorphisms in high dimension are ``wild objects,'' and, even with the kind of regularization that is applied in this paper, it was not obvious that using them in a non-parametric setting would avoid overfitting and compare, often favorably, with several state-of-the-art machine learning methods.
%Compared even to neural networks, diffeomorphic transformations are indeed very high dimensional objects. Because they are generated as flows associated with ordinary differential equations (ODEs), they require a time step discretization, and at each time step the estimation of a vector field providing the right-hand side of the ODE. Our current ``exact'' parametrization of the problem involves $TNd$ parameters where $T$ is the number of time steps in the discretization, $N$ is the number of training samples and $d$ the dimension of the data, a number that can easily become extremely large (some strategies are described in section \ref{sec:dim} in order to design approximations). 
As explained in section \ref{sec:dummy}, the model (after the addition a  ``dummy'' dimension) can represent essentially any function of the data. 

In this regard, given that one is ready to work with such high-dimensional objects, the requirement that the transformation is a diffeomorphism is not a strong restriction to the scope of the model. Using such a representation has several advantages, however. It indeed expresses the model in terms of a well-understood non-linear space of functions (the group of diffeomorphisms), which has a simple algebraic structure, and a differential geometry extensively explored in mathematics. The description of such models is, as a consequence, quite concise and explicit, and lends itself to an optimal control formulation with the learning algorithms that result from it. Finally, there are obvious advantages, on the generative side, in using invertible data transformations, especially when the data in the transformed configuration may be easier to describe using a simple statistical model, since, in that case, the application of the inverse diffeomorphism to realizations of that simple model provides a generative model in the original configuration space.

The paper is organized as follows. Section \ref{sec:distance} introduces basic concepts related to groups of diffeomorphisms and their Riemannian distances. Section \ref{sec:opt.cont} formulates the optimal control problem associated to the estimation of the diffeomorphic predictor, introducing its reduction to a finite-dimensional parametrization using a kernel trick and describing its optimality conditions. Some additional discussion on the reproducing kernel is provided in section \ref{sec:kernel}.   Section \ref{sec:remarks} introduces remarks and extensions that are specific to the prediction problems that we consider here, and sections \ref{sec:experiments} and \ref{sec:param} provide experimental results, including details on how (hyper-)parameters used in our model can be set. We conclude the paper in section \ref{sec:discussion}.

\section{Distance on Diffeomorphisms}
\label{sec:distance}

We now describe the basic framework leading to the definition of Riemannian metrics on groups of diffeomorphisms. While many of the concepts described in these first sections are adapted from similar constructions in shape analysis \citep{younes2010shapes}  this short presentation helps making the paper self-contained and accessible to readers without prior knowledge of that domain.

Let $\mathbf B_p = C^p_0(\mR^d, \mR^d)$ denote the space of $p$-times continuously differentiable functions that tend to 0 (together with their first $p$ derivatives) to infinity. This space is a Banach space, for the norm
\[
\|v\|_{p,\infty} = \max_{0\leq k \leq p}   \|d^kv\|_\infty
\] 
where $\|\cdot\|_\infty$ is the usual supremum norm.

Introduce a Hilbert space $V$ of vector fields on $\mR^d$ which is continuously embedded in $\mathbf B_p$ for some $p\geq 1$, which means that there exists a constant $C$ such that
\[
 \|v\|_{p,\infty} \leq C \|v\|_V
 \]
 for all $v$ in $V$, where $\|\cdot\|_V$ denotes the Hilbert norm on $V$ (and we will denote the associated inner product by $\scp{\cdot}{\cdot}_V$). This assumption implies that $V$ is a reproducing kernel Hilbert space (RKHS). Because $V$ is a space of vector fields, the definition of the associated kernel slightly differs from the usual case of scalar valued functions in that the kernel is matrix valued. More precisely, a direct application of Riesz's Hilbert space representation theorem implies that there exists a function
 \[
 K: \mR^d \times \mR^d \to \mathcal M_d(\mR)
 \]
 where $\mathcal M_d(\mR)$ is the space of $d$ by $d$ real matrices, such that
 \begin{enumerate}
 \item The vector field $K(\cdot, y)a: x \mapsto K(x,y)a$ belongs to $V$ for all $y, a\in \mR^d$.
 \item For $v\in V$, for all $y,a\in \mR^d$, $\scp{K(\cdot, y)a}{v}_V = a^T v(y)$.
 \end{enumerate}
 These properties imply that $\scp{K(\cdot, x)a}{K(\cdot, y)b}_V = a^T K(x,y) b$ for all $x,y,a,b\in \mR^d$, which in turn implies that $K(y,x) = K(x,y)^T$ for all $x, y\in \mR^d$. \\
 
 Diffeomorphisms can be generated as flows of ordinary differential equations (ODEs) associated with time-dependent elements of $V$. More precisely, let $v \in L^2([0,1], V)$, i.e., $v(t)\in V$ for $t \in [0,1]$ and
 \[
 \int_0^1 \|v(t)\|_V^2\,dt < \infty.
 \] 
 Then, the ODE $\prt_t y = v(t, y)$ has a unique solution over $[0,1]$, and the flow associated with this ODE is the function $\phi^v: (t, x) \mapsto y(t)$ where $y$ is the solution starting at $x$. This flow is, at all times, a diffeomorphism of $\mR^d$, and satisfies the equation $\prt_t \phi = v(t)\circ \phi$, $\phi(0) = \id$ (the identity map).  Here, and in the rest of this paper, we make the small abuse of notation of writing $v(t)(y) = v(t,y)$, where, in this case, $v(t)\in V$ for all $t\in [0,1]$.  Similarly, we will often write $\phi^v(t)$ for the function $x \mapsto \phi^v(t, x)$, so that $\phi^v$ may be considered either as a time-dependent diffeomorphism, or as a function of both time and space variables. 
 \medskip
 
The set of diffeomorphisms that can be generated in such a way forms a group, denoted $\Diff_V$ since it depends on $V$. Given $\psi_1\in \Diff_V$, one defines the optimal deformation  cost $\Lambda(\psi_1)$ from $\id$ to $\psi_1$ as the minimum of 
 $\int_0^1 \|v(t)\|_V^2\,dt$ over all $v\in L^2([0,1], V)$ such that $\phi^v(1) = \psi_1$. If we let $D(\psi_1, \psi_2) = \Lambda(\psi_2\circ \psi_1^{-1})^{1/2}$ then $D$ is a geodesic distance on $\Diff_V$ associated with the right-invariant Riemannian metric generated by $v\mapsto \|v\|_V$ on $V$. We refer to \cite{younes2010shapes} for more details and additional properties on this construction. For our purposes here, we only need to notice that the minimization of the objective function in \eqref{eq:cost} can be rewritten as an optimal control problem minimizing
\begin{equation}
\label{eq:cost.opt}
E(v, \theta) = \int_0^1 \|v(t)\|_V^2\, dt + \lambda g(\tht) + \frac1{\sigma^2} \Gamma(F(\cdot, \theta), \phi(1)\cdot \mathcal T_0)
\end{equation}
over $v\in L^2([0,1], V)$, $\theta\in \mR^q$ and subject to the constraint that $\phi(t)$ satisfies the equation $\prt_t \phi = v\circ \phi$ with $\phi(0) = \id$. We point out that, under mild regularity conditions on the dependency of $\Ga$ with respect to $\mathcal T_0$ (continuity in $x_1, \ldots, x_N$ suffices), a minimizer of this function in $v$ for fixed $\tht$ always exists, with $v \in L^2([0,1], V)$. 

\section{Optimal Control Problem}
\label{sec:opt.cont}
\subsection{Reduction}
\label{sec:reduction}
The minimization in \eqref{eq:cost.opt} can be reduced using an RKHS argument, similar to the kernel trick invoked in standard kernel methods \citep{aro50,duc76,mei79,wah90,ss02}. Let $z_k(t) = \phi(t, x_k)$. Because the endpoint cost $\Gamma$ only depends on $(z_1(1), \ldots, z_N(1))$, it only suffices to compute these trajectories, which satisfy $\prt_t z_k = v(t, z_k)$. This implies that an optimal $v$ must be such that, at every time $t$, $\|v(t)\|_V^2$ is minimal over all $\|w\|_V^2$ with $w$ satisfying $w(z_k) = v(t, z_k)$, which requires $v(t)$ to take the form
\begin{equation}
\label{eq:v.reduc}
v(t, \cdot) = \sum_{k=1}^N K(\cdot, z_k(t)) a_k(t)
\end{equation}
where $K$ is the kernel of the RKHS $V$ and $a_1, \ldots, a_N$ are unknown time-dependent vectors in $\mR^d$, which provide our reduced variables. Letting $\bsa  = (a_1, \ldots, a_N)$, the reduced problem requires to minimize
\begin{equation}
\label{eq:cost.red}
E(\bsa(\cdot), \theta) = \int_0^1 \sum_{k,l=1}^N a_k(t)^TK(z_k(t), z_l(t)) a_l(t) \, dt + \lambda g(\theta) + \frac1{\sigma} \Gamma(F(\cdot, \theta), \mathcal T(\bsz(1)))
\end{equation}
subject to $\prt_t z_k = \sum_{l=1}^N K(z_k, z_l) a_l$, $z_k(0) = x_k$, with the notation $\bsz = (z_1, \ldots, z_N)$ and $\mathcal T(\bsz) = (z_1, y_1, \ldots, z_N, y_N)$. 
 
\subsection{Optimality Conditions and Gradient}
\label{sec:grad}
We now consider the minimization problem with fixed $\tht$ (optimization in $\tht$ will depend on the specific choice of classifier $F$ and risk function $\Ga$). For the optimal control problem \eqref{eq:cost.red}, the ``state space'' is the space in which the ``state variable'' $\bsz = (z_1, \ldots, z_N)$ belongs, and is therefore $Q = (\mR^d)^N$. The control space contains the control variable $\bsa$, and is $U = (\mR^d)^N$.
%(even though these spaces are identical, it is conceptually useful to keep separate notation for both). 
  
Optimality conditions for the variable $\bsa$ are provided by Pontryagin's maximum principle (PMP). They require the introduction of a third variable (co-state), denoted $\bsp\in Q$, and of a control-dependent Hamiltonian $H_\bsa$ defined on $Q\times Q$ given, in our case, by
\begin{equation}
\label{eq:ham}
H_\bsa(\bsp, \bsz) = \sum_{k,l=1}^N (p_k-a_k)^TK(z_k, z_l) a_l.
\end{equation}
(In this expression, $\bsa$, $\bsp$ and $\bsz$ do not depend on time.) The PMP \citep{hocking1991optimal,macki2012introduction,miller2015hamiltonian,vg97} then states that any optimal solution $\bsa$ must be such that there exists a time-dependent co-state satisfying
\begin{equation}
\label{eq:pmp}
\left\{
\begin{aligned}
&\prt_t \bsz = \prt_{\bsp} H_{\bsa(t)}(\bsp(t), \bsz(t))\\
&\prt_t \bsp = -\prt_{\bsz} H_{\bsa(t)}(\bsp(t), \bsz(t))\\
&\bsa(t) = \mathrm{argmax}_{\bsa'} H_{\bsa'}(\bsp(t), \bsz(t))\\
 \end{aligned}
 \right.
 \end{equation}
 with boundary conditions $\bsz(0) = (x_1, \ldots, x_N)$ and
 \begin{equation}
 \label{eq:pmp.end}
 \bsp(1) = - \frac1{\sigma^2} \prt_{\bsz} \Gamma(F(\cdot, \theta), \mathcal T(\bsz(1))).
 \end{equation}
 
 These conditions are closely related to those allowing for the computation of the differential of $E$ with respect to $\bsa(\cdot)$, which is given by
 \[
 \prt_{\bsa(\cdot)} E(\bsa(\cdot), \theta) = \bsu(\cdot)
 \]
 with
 \begin{equation}
 \label{eq:diff}
 u_k(t) = \sum_{l=1}^N K(z_k(t), z_l(t)) (p_l(t) - 2a_l(t))
 \end{equation}
 where $p$ solves
 \begin{equation}
\label{eq:pmp.grad}
\left\{
\begin{aligned}
&\prt_t \bsz = \prt_{\bsp} H_\bsa(t)(\bsp(t), \bsz(t))\\
&\prt_t \bsp = -\prt_{\bsz} H_\bsa(t)(\bsp(t), \bsz(t))
 \end{aligned}
 \right.
 \end{equation}
 with boundary conditions $\bsz(0) = (x_1, \ldots, x_N)$ and
 \[
 \bsp(1) = - \frac{1}{\sigma^2} \prt_{\bsz} \Gamma(F(\cdot, \theta), \mathcal T(\bsz(1))).
 \]
 Concretely, the differential is computed by (i) solving the first equation of \eqref{eq:pmp.grad}, which does not involve $\bsp$, (ii) using the obtained value of $\bsz(1)$ to compute $\bsp(1)$ from the boundary condition,  then (iii) solving the second equation of \eqref{eq:pmp.grad} backward in time to obtain $\bsp$ at all times, and (iv) plug it in the definition of $\bsu(t)$. \\

For practical purposes, the discrete time version of the problem is obviously more useful, and its differential is obtained from a similar dynamic programming (or back-propagation) computation. Namely, discretize time over $0, 1, \ldots, T$ and consider the objective function 
\begin{equation}
\label{eq:cost.red.2}
E(\bsa(\cdot), \theta) = \frac1T \sum_{t=0}^{T-1} \sum_{k,l=1}^N a_k(t)^TK(z_k(t), z_l(t)) a_l(t) \, dt + \lambda g(\tht) + \frac1{\sigma^2} \Gamma(F(\cdot, \theta), \mathcal T(\bsz(T)))
\end{equation}
subject to 
\[
z_k(t+1) = z_k(t) + \frac1T \sum_{l=1}^N K(z_k(t), z_l(t)) a_l(t),\quad z_k(0) = x_k.
\]
 We therefore use a simple Euler scheme to discretize the state ODE. Note that the state is discretized over $0, \dots, T$ and the control over $0, \dots, T-1$. 
The differential of $E$ is now given by the   following expression, very similar to that obtained in continuous time.
 \[
 \prt_{\bsa(\cdot)} E(\bsa(\cdot), \theta) = \bsu(\cdot)
 \]
 with
 \[
 u_k(t) = \sum_{l=1}^N K(z_k(t), z_l(t)) (p_l(t) - 2a_l(t)), \quad t=0, \ldots, T-1
 \]
 where $p$ (discretized over $0, \ldots, T-1$), can be computed using
 \begin{equation}
\label{eq:pmp.grad.2}
\left\{
\begin{aligned}
&\bsz(t+1) = \bsz(t) + \frac1T \prt_{\bsp} H_\bsa(t)(\bsp(t), \bsz(t))\\
&\bsp(t-1) = \bsp(t) + \frac1T \prt_{\bsz} H_\bsa(t)(\bsp(t), \bsz(t))
 \end{aligned}
 \right.
 \end{equation}
 with boundary conditions $\bsz(0) = (x_1, \ldots, x_N)$ and
 \[
 \bsp(T-1) = - \frac1{\sigma^2} \prt_{\bsz} \Gamma(F(\cdot, \theta), \mathcal T(\bsz(T))).
 \]
 These computations allow us to compute the differential of the objective function with respect to $\bsa$. The differential in $\theta$ depends on the selected terminal classifier and its expression for the function chosen in \eqref{eq:Log.reg} is standard.

We emphasize the fact that we are talking of differential of the objective function rather than its gradient. Our implementation uses a Riemannian (sometimes called ``natural'') gradient with respect to the metric
\[
\scp{\eta_1(\cdot)}{\eta_2(\cdot)}_{\bsa(\cdot)} = \int_0^1 \sum_{k,l=1}^n \eta_k(t)^T K(z_k(t), z_l(t)) \eta_l(t) dt
\]
with $\prt_t z_k =  \sum_{l=1}^n K(z_k(t), z_l(t)) a_l(t)$. With respect to this metric, one has
\[
 \nabla_{\bsa(\cdot)} E(\bsa(\cdot), \theta) = \bsp - 2\bsa,
\]
a very simple expression that can also be used in the discrete case. Using this Riemannian inner product as a conditioner for the deformation parameters (and a standard Euclidean inner product  on the other parameters), experiments in sections \ref{sec:experiments} and \ref{sec:param} run Polak-Ribiere conjugate gradient iterations \citep{nw99} to optimize the objective function. (We also experimented with limited-memory BFGS, which does not use natural gradients,  and found that conditioned conjugate gradient performed better on our data.) 

\section{Kernel}
\label{sec:kernel}
\subsection{General Principles}
To fully specify the algorithm, one needs to select the RKHS $V$, or, equivalently, its reproducing kernel, $K$.   They constitute important components of the model because they determine the regularity of the estimated diffeomorphisms. We recall that $K$ is a kernel over vector fields, and therefore is matrix valued. One simple way to build such a kernel is to start with a scalar positive kernel $\kappa: \mR^d \times \mR^d \to \mR$ and let
\begin{equation}
\label{eq:scalar.k}
K(x,y) = \kappa(x,y) \mathrm{Id}_{\mR^d}.
\end{equation}
We will refer to such kernels as ``scalar.''

One can choose $\kappa$ from the large collection of known positive kernels (and their infinite number of possible combinations; \cite{aro50,dyn89,ss02,buh03}). Most common options are Gaussian kernels,
\begin{equation}
\label{eq:gauss.k}
\ka(x,y) = \exp(- |x-y|^2/2\rho^2),
\end{equation}
or Mat\'ern kernels of class $C^k$,
\begin{equation}
\label{eq:matern.k}
\ka(x,y) = P_k(|x-y|/\rho)\exp(-|x-y|/\rho),
\end{equation}
where $P_k$ is a reversed Bessel polynomial of order $k$. In both cases, $\rho$ is a positive scale parameter. The Mat\'ern kernels have the nice property that their associated RKHS is equivalent to a Sobolev space of order $k + d/2$. 

Vector fields $v$ in the  RKHS associated with a matrix kernel such as \eqref{eq:scalar.k}, where $\ka$ is a radial basis function (RBF), are such that each coordinate function of $v$ belongs to the scalar RKHS associated with $\ka$, which is translation and rotation invariant (i.e., the transformations that associate to a scalar function $h$ the functions $x \mapsto h(R^T(x-b))$ are isometries, for all rotation matrices $R$ and all vectors $b\in \mR^d$).

\subsection{Graph-Based Kernels}

While \eqref{eq:scalar.k} provide a simple recipe for the definition of matrix-valued kernels, other interesting choices can be made within this class. Rotation and translation invariance more adapted to spaces of vector fields, in which one requires that replacing $v: \mR^d \to \mR^d$ by $x \mapsto Rv(R^T(x-b))$ is an isometry of $V$ for all $R,b$, leads to a more general class of matrix kernels extensively discussed in  \citep{Micheli_TRIKernels_2014}. When the data is structured, however (e.g., when it is defined over a grid), rotation invariance may not be a good requirement since it allows, for example, for permuting coordinates, which would break the data structure. In this context, other choices may  be preferable, as illustrated by the following example.  Assume that the data is defined over a graph, say $\mathcal G$ with $d$ vertices. Then, one may consider matrix kernels relying on this structure. For example, letting $\mathcal N_i$ denote the set of nearest neighbors of $i$ in $\mathcal G$, one can simply take
\begin{equation}
\label{eq:kernel.nn}
K(x,y) = \mathrm{diag}(\Phi(|P_ix - P_iy|), i = 1, \ldots, d)
\end{equation} 
where $P_ix$ is the vector $(x_j, j\in \mathcal N_i)$ and $\Phi$ is an RBF associated to a positive radial scalar kernel.

\subsection{Introducing Affine Transformations}
\label{sec:affine}
RKHS's of vector fields built from RBF's have the property that all their elements (and several of their derivatives) vanish at infinity. As a consequence, these spaces do not contain simple transformations, such as translations or more general affine transformations. It is however possible to complement them with such mappings, defining
\[
\hat V_{\mathfrak a} = \{ g+v: g\in \mathfrak a, v\in V\}
\]
where $\mathfrak a$ is any Lie sub-algebra of the group of affine transformations (so that any element $g\in \mathfrak a$ takes the form $g(x) = Ax+b$, where $A$ is a matrix and $b$ is a vector). In particular, $\mathfrak a$ can be the whole space of affine transformations. Since $\mathfrak a$ and $v$ intersect only at $\{0\}$, one can define without ambiguity a Hilbert norm on $\hat V_{\mathfrak a}$ by letting
\[
\|g+v\|^2_{\hat V_{\mathfrak a}} = \|g\|_{\mathfrak a}^2 + \|v\|_V^2
\]
where $\|g\|_{\mathfrak a}$ is any inner-product norm on $\mathfrak a$. A simple choice, for $g(x) = Ax + b$, can be to take
\begin{equation}
\label{eq:aff.norm}
\|g\|_{\mathfrak a}^2 = \ka_1 \mathrm{trace}(AA^T) + \ka_2 |b|^2.
\end{equation}
Both $\mathfrak a$ and $\hat V_{\mathfrak a}$ are RKHS's in this case, respectively with kernels $K_{\mathfrak a}$ and $K_{\mathfrak a} + K$, where $K$ is the kernel of $V$.   
If the norm on $\mathfrak a$ is given by \eqref{eq:aff.norm}, then
\[
K_{\mathfrak a}(x,y) = \left(\frac{x^Ty}{\ka_1} + \frac1{\ka_2}\right) \mathrm{Id}_{\mR^d}, 
\] 
as can be deduced from the definition of a reproducing kernel.
\bigskip

Instead of using this extended RKHS, one may prefer to model  affine transformations separately from the vector field. This leads to replacing \eqref{eq:v.reduc} by
\[
\hat v(t, \cdot) = g(t, \cdot) + \sum_{k=1}^N K(\cdot, z_k(t)) a_k(t)
\]
and the cost \eqref{eq:cost.red} by 
\[
E(\bsa(\cdot), \theta) = \int_0^1 \|g(t)\|^2_{\mathfrak a} \, dt + \int_0^1 \sum_{k,l=1}^N a_k(t)^TK(z_k(t), z_l(t)) a_l(t) \, dt + \lambda g(\tht) +\frac1{\sigma^2} \Gamma(F(\cdot, \theta), \mathcal T(1))
\]
with $\prt_t z_k = \hat v(t, z_k(t))$, $z_k(0) = x_k$. The two approaches are, in theory, equivalent, in that the second one simply ignore the reduction on the affine part of the vector field, but it may be helpful, numerically, to use separate variables in the optimization process for the affine transform and the vector field reduced coefficients, because this gives more flexibility to the optimization. The derivation of the associated optimality conditions and gradient are similar to those made in section \ref{sec:grad} and left to the reader. 
\bigskip

Another point worth mentioning is that, if $\phi$ satisfies 
\[
\prt_t \phi = g\circ \phi + v\circ \phi
\]
for (time-dependent) $g\in \mathfrak a$ and $v\in V$, and if one defines the time-dependent affine transformation $\rho$ by
\[
\prt_t \rho = g \circ \rho,
\]
then $\phi = \rho\circ \psi$, where $\psi$ satisfies $\prt_t \psi = w \circ \psi$ and $w = \rho_L^{-1} v\circ \rho$, $\rho_L$ being the linear part of  $\rho$. In the special case when $\mathfrak a$ is the Lie algebra of the Euclidean group, so that $\rho$ is the composition of a rotation and a translation,  and when the norm of $V$ is invariant by such transformations (e.g., when using a scalar kernel associated to an RBF), then $\phi$ and $\psi$ are equidistant to the identity. As a consequence, when the final function $F$ implements a linear model, there is, in theory (numerics may be different) no gain in introducing an affine component restricted to rotations and translations. The equidistance property does not hold, however, if one uses a larger group of affine transformations, or a norm on $V$ that is not Euclidean invariant, and the introduction of such transformations actually extends the model in a way that may  significantly modify its performance, generally, in our experiments, for the better. 
`

 \section{Enhancements and Remarks}
 \label{sec:remarks}
 
 \subsection{Adding a Dimension}
 \label{sec:dummy}
 We use, in this paper, logistic regression as final classifier applied to transformed data. This classifier estimates a linear separation rule  between the classes, but it should be clear that not every training set can be transformed into a linearly separable one using a diffeomorphism of the ambient space. A very simple one-dimensional example is when the true class associated to an input $x\in \mR$ is 0 if $|x|<1$ and 1 otherwise: no one-dimensional diffeomorphism will make the data separable, since such diffeomorphisms are necessarily monotone. This limitation can be fixed easily, however, by adding a dummy dimension and apply the model to a $(d+1)$-dimensional dataset in which $X$ is replaced by $(X,0)$. In the example just mentioned, for example, the transformation $(x,\mu) \mapsto (x, \mu+ x^2-1)$ is a diffeomorphism of $\mR^2$ that separates the two classes (along the $y$ axis).  
 
\begin{figure}[h]
 \centering
 \includegraphics[trim= 2cm 8cm 2cm 2cm, clip, width=0.35\textwidth]{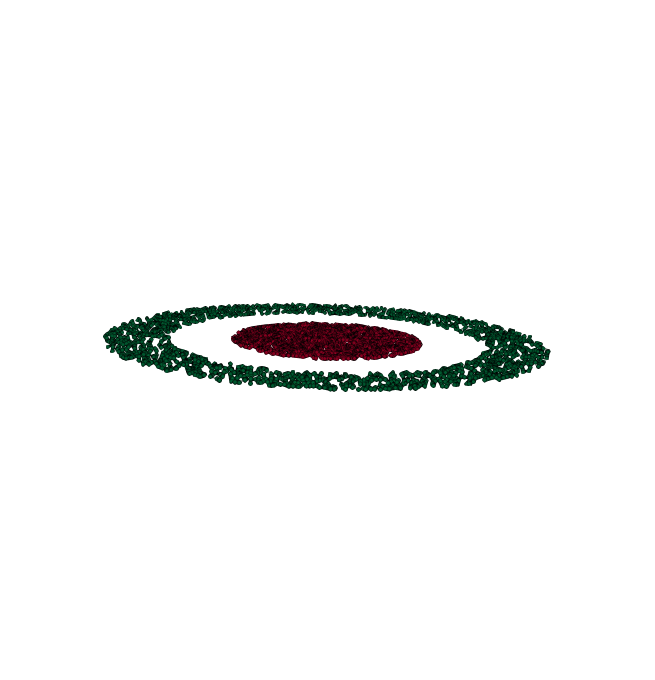}
 \includegraphics[trim= 2cm 8cm 2cm 2cm, clip, width=0.35\textwidth]{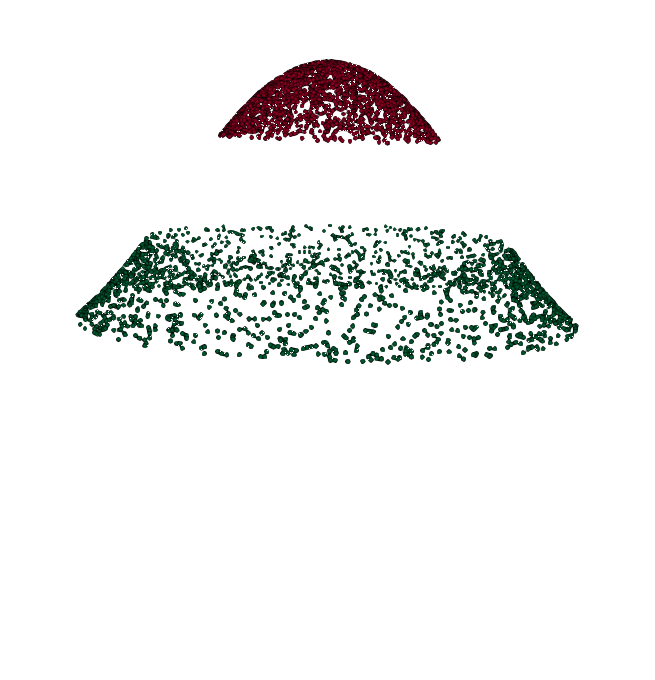}
 \caption{ \label{fig:odot} Additional dimension separating the center of a target from its external ring. Left: initial configuration; Right: transformed configuration.}
 \end{figure}

\begin{figure}[h]
 \centering
 \includegraphics[trim= 0cm 6cm 0cm 8cm, clip, width=0.35\textwidth]{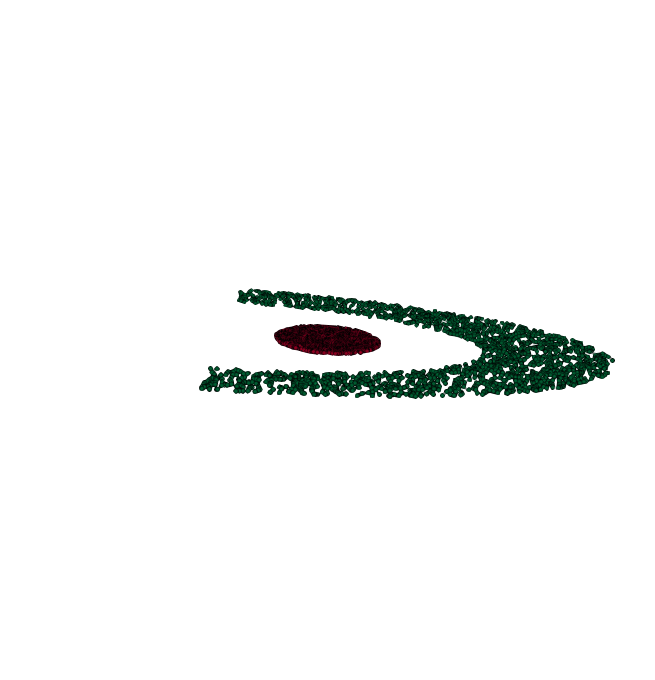}
 \includegraphics[trim= 0cm 6cm 0cm 8cm, clip, width=0.35\textwidth]{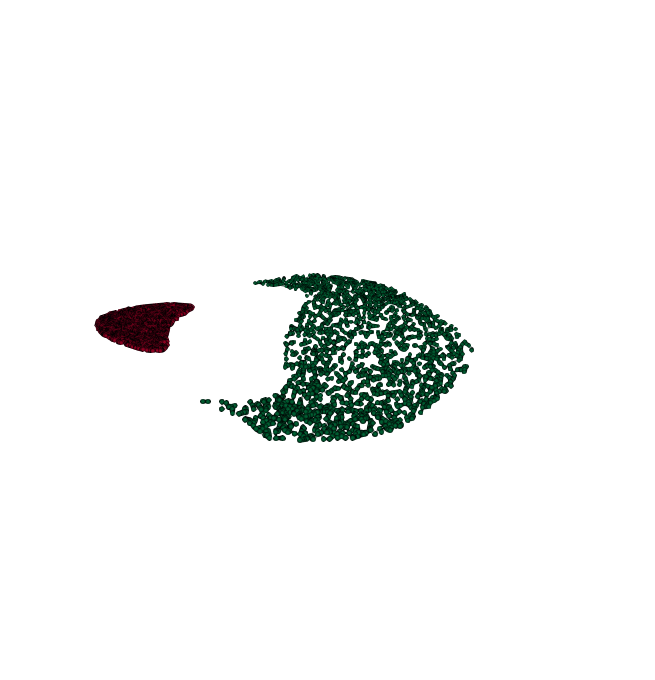}\\
 \includegraphics[trim= 0cm 6cm 0cm 4cm, clip, width=0.35\textwidth]{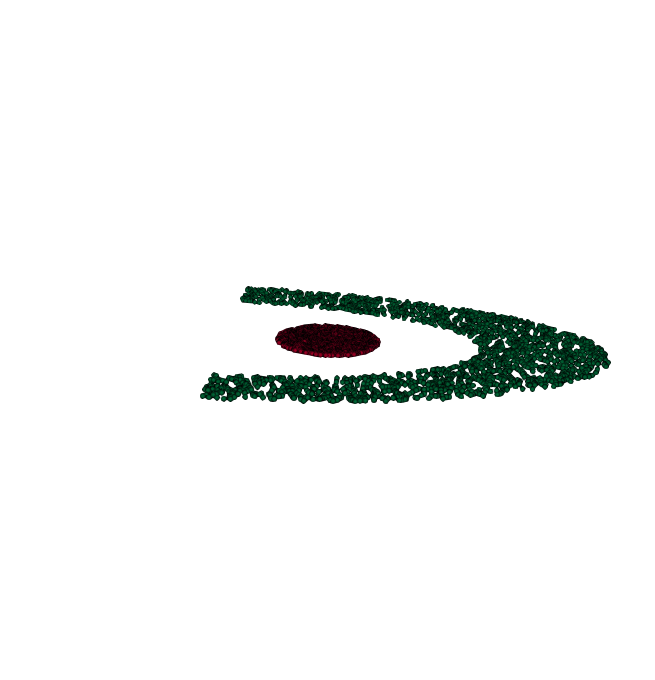}
 \includegraphics[trim= 0cm 6cm 0cm 4cm, clip, width=0.35\textwidth]{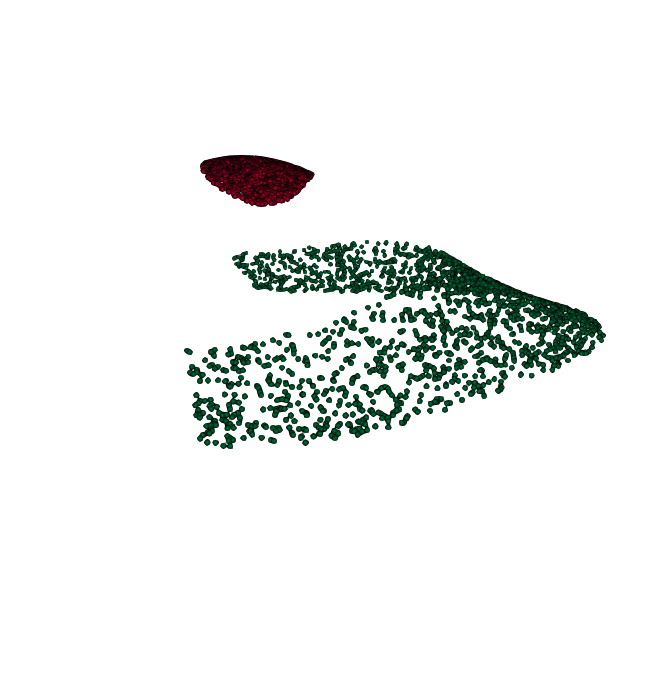}
 \caption{ \label{fig:cdot} Comparison of  optimal transformations without (top) and with (bottom) adding a dimension. Left: initial configuration; Right: transformed configuration.}
 \end{figure}

 Notice that any binary classifier that can be expressed as $x \mapsto \mathrm{sign}(f(x) - a)$ for some smooth function $f$ can be included in the model class we are considering after adding a dimension, simply taking (letting again $\mu$ denote the additional scalar variable) $\psi(x,\mu) = (x, \mu + f(x))$, which is a diffeomorphism of $\mR^{d+1}$, and $u = (0_{\mR^d}, 1)$. However, even when it works perfectly on the training set, this transformation will  not be optimal in general, and the diffeomorphic classifier would typically prefer a diffeomorphism $\phi$ that will minimize the overall distortion, essentially trading off some non-linear transformation of the data, $x$, to induce a ``simpler'' classification rule. Figure \ref{fig:odot} provides an example of the effect of adding a dimension when the original data is two-dimensional with two classes forming a target pattern. While no 2D transformation will break the topology and separate the central disk from the exterior ring, using an additional dimension offers a straightforward solution. Another simple example is provided in Figure \ref{fig:cdot}, where a circle (class 1) is inscribed in a half ellipse (class 2). In this case, adding a dimension is not required, but the transformation estimated when this is done is closer to the identity (in the sense of our metric on diffeomorphisms) than the one computed in two dimensions.
 
 When adding one or more dimensions (in a $c$-class problem, it makes sense to add $c-1$ dimensions), one may want to expand  training data in the form   $x_k \to (x_k, \delta u_k)$ for small $\delta$, with $u_k$ a realization of a standard Gaussian variable  to help breaking symmetries in the training phase (test data  being still expanded as  $x_k \to (x_k, 0)$).

\subsection{Parametric Dimension and Sub-Riemannian Methods}
\label{sec:dim}

The dimensional reduction method described in section \ref{sec:reduction} is the exact finite-dimensional parametrization of the spatial component of the original infinite-dimensional problem. Assuming $T$  steps for the time discretization, this results in $TdN$ parameters in the transformation part of the model, while the computation of the gradient has a $TdN^2$ order cost. With our implementation (on a four-core Intel i7 laptop), we were able to handle up to 2,000 training samples in 100 dimensions with 10 time steps, which required about one day for 2,000 gradient iterations. Even if the efficiency of our implementation may be improved, for example through the use of GPU arrays, it is clear that the algorithm we just described will not scale to large datasets unless some approximations are made. Importantly, no significant reduction in computation cost can be obtained by randomly sampling from the training set at each iteration, since the representation \eqref{eq:v.reduc} requires computing the trajectories of all sample points. 

This suggests  replacing the optimal expression in \eqref{eq:v.reduc} by a form that does not requires tracking the trajectories of $N$ points. One possibility is to require that $v$ is decomposed as a sum similar to \eqref{eq:v.reduc}, but over a smaller number of ``control points'', i.e., to let
\begin{equation}
\label{eq:v.reduc.2}
v(t, \cdot) = \sum_{j=1}^n K(\cdot, \zeta_{j}(t)) a_j(t)
\end{equation}
with $\prt_t \zeta_{j} = v(t, \zeta_{j})$ and $\zeta_j(0) = \zeta^0_{j}$ where $\{\zeta^0_1, \ldots, \zeta^0_n\}$ is, for example,  a subset of the training data set. 
%The computation of an exact gradient using now has $TdnN$ complexity, and the $N$ factor can further be reduced using stochastic gradient strategies, which become possible because the expression of the vector field is now decoupled from the training set.
As a result, the computational cost per iteration in the data size would now be of order $TdNn$ (because the evolution of all points is still needed to evaluate the objective function and its gradient), but this change would make possible randomized evaluations of the data cost, $\Gamma$, (leading to stochastic gradient descent) that would replace $TdnN$ by $TdnN'$ where $N'$ is the size of the mini-batch used at each iteration. This general scheme will be explored in future work,  where optimal strategies in selecting an $n$-dimensional subset of the training data must be analyzed, including in particular the trade-off they imply between  computation time and sub-optimality of solutions.  Even though they appeared in a very different context, some inspiration may be obtained from similar approaches that were explored in shape analysis \citep{younes2012constrained,durrleman2013sparse,younes2014gaussian,durrleman2014morphometry,gris2018sub}.

Another plausible choice is to specify a parametric form for the vector field at a given time. It is quite appealing to use a neural-net-like expression, letting
\[
v(t, x) = \Phi(A(t) x + b(t))
\]
where $A$ is a time-dependent $d$ by $d$ matrix, $b$ a time dependent vector and $\Phi$ a function applying a fixed nonlinear transformation  (for example, $t \mapsto t \exp(-t^2/2)$) to each coordinate. While the exact expression of $\|v(t)\|_V$ may be challenging to compute analytically when $v$ is given in this form, it is sufficient, in order to ensure the existence of solutions to the state equation, to control the supremum norm of the first derivative of $v(t)$, which, since $\Phi$ is fixed, only depends of $A(t)$. One can therefore replace $\|v(t)\|_2$ in the optimal control formulation \eqref{eq:cost.opt} by any norm evaluated at $A(t)$ in  the finite-dimensional space of $d$ by $d$ matrices. Preliminary experiments, run on some of the examples discussed in section \ref{sec:experiments}, show that such a representation can perform very well in some cases and completely fail in others, clearly requiring further study to be reported in future work. The computation time is, in all cases, significantly reduced compared to the original version.

 \subsection{Deformable Templates}

 It is important to strengthen the fact that, even though our model involves diffeomorphic transformations, it is not a deformable template model. The latter type of model typically works with small-dimensional images (k=2 or 3), say $I: \mR^k\to \mR$, and tries to adjust a $k$-dimensional deformation (using a diffeomorphism $g: \mR^k \to \mR^k$) such that the deformed image, given by $I\circ g^{-1}$ aligns with a fixed template (and classification based on a finite number of templates would pick the one for which a combination of the deformation and the associated residuals is smallest).
 
 The transformation $\psi_g: I \mapsto I \circ g^{-1}$ is a homeomorphism of the space of, say, continuous images. Once images are discretized over a grid with $d$ points, $\psi_g$ becomes (assuming that the grid is fine enough) a one-to-one transformation of $\mR^d$, but a very special one. In this context, the model described in this paper would be directly applied to discrete images, looking for a $d$-dimensional diffeomorphism that would include deformations such as $\psi_g$, but many others, involving also variations in the image values or more complex transformations (including, for example, reshuffling all image pixels in arbitrary orders!).

\section{Comparative Experiments}
\label{sec:experiments}

\subsection{Classifiers Used for Comparison}
\label{sec:classif}
We now provide a few experiments that illustrate some of the advantages of the proposed diffeomorphic learning method, and some of its limitations as well. We will compare the performance of this algorithm with a few off-the-shelve methods, namely $k$-nearest-neighbors ($k$NN), linear and non-linear SVM, random forests (RF), multi-layer perceptrons with 1, 2 and 5 hidden layers (abbreviated below as MLP1, MLP2 and MLP5) and logistic regression \citep{htf03,bishop2006pattern}. The classification rates that were reported were evaluated on a test set containing 2,000 examples per class (except for MNIST, for which we used the test set available with this data).  

We used the scikit-learn Python package \citep{pedregosa2011scikit}  with the following parameters (most being default in scikit-learn). \begin{enumerate}[label=$\bullet$]
\item Linear SVM: $\ell^2$ penalty with default weight $C=1$, with one-vs-all multi-class strategy when relevant.
\item Kernel SVM: $\ell^2$ penalty with weight $C$ estimated as described in section \ref{sec:param}. Gaussian (i.e., RBF) kernel with coefficient $\gamma$ identical to that used for for the kernel in diffeomorphic learning. One-vs-all multi-class strategy when relevant.
\item Random forests:  100 trees, with Gini entropy splitting rule, with the default choice ($\sqrt d$) of number of features at each node.
\item $k$-nearest neighbors: with the standard Euclidean metric and the default (5) number of neighbors.
\item Multi-layer perceptrons: with ReLU activations, ADAM solver, constant learning rate and 10,000 maximal iterations, using 1, 2 or 5 hidden layers each composed of 100 units.
\item Logistic regression: $\ell^2$ penalty with weight $C=1$. The same classifier is used as the final step of the diffeomorphic learning algorithm, so that its performance on transformed data is also the performance of the algorithm that is proposed in this paper.
\end{enumerate} 

In all cases, we added a dummy dimension to the data as described in section \ref{sec:dummy} when running diffeomorphic learning (classification results on original data were obtained without the added  dimension). We also used an affine transformation to complement the kernel, as described in section \ref{sec:affine}. Note that adding a dimension was not always necessary for learning (especially with large dimensional problems), nor was the affine transform always improving on results, but they never harm the results in any significant way, so that it was simpler to always turn on these options in our experiments. The optimization  procedure was initialized with a vanishing vector field (i.e., $\psi = \id$) and run until numerical stabilization (with a limit of 2,000 iterations at most). Doing so was always an overkill, in terms of classification performance, because in almost all cases, the limit classification error on the test set stabilizes faster than the time taken by the algorithm to optimize the transformation. It was however important to avoid stopping the minimization early in order to make sure that optimizing the diffeomorphism did not result in overfitting. 

We now describe the datasets that we used (all but the last one being synthetic) and compare the performances of the classifiers above on the original data and on the transformed data after learning.

\subsection{Tori datasets}
\begin{figure}[h]
\centering
\includegraphics[trim=3cm 3cm 3cm 3cm, clip, width=0.3\textwidth]{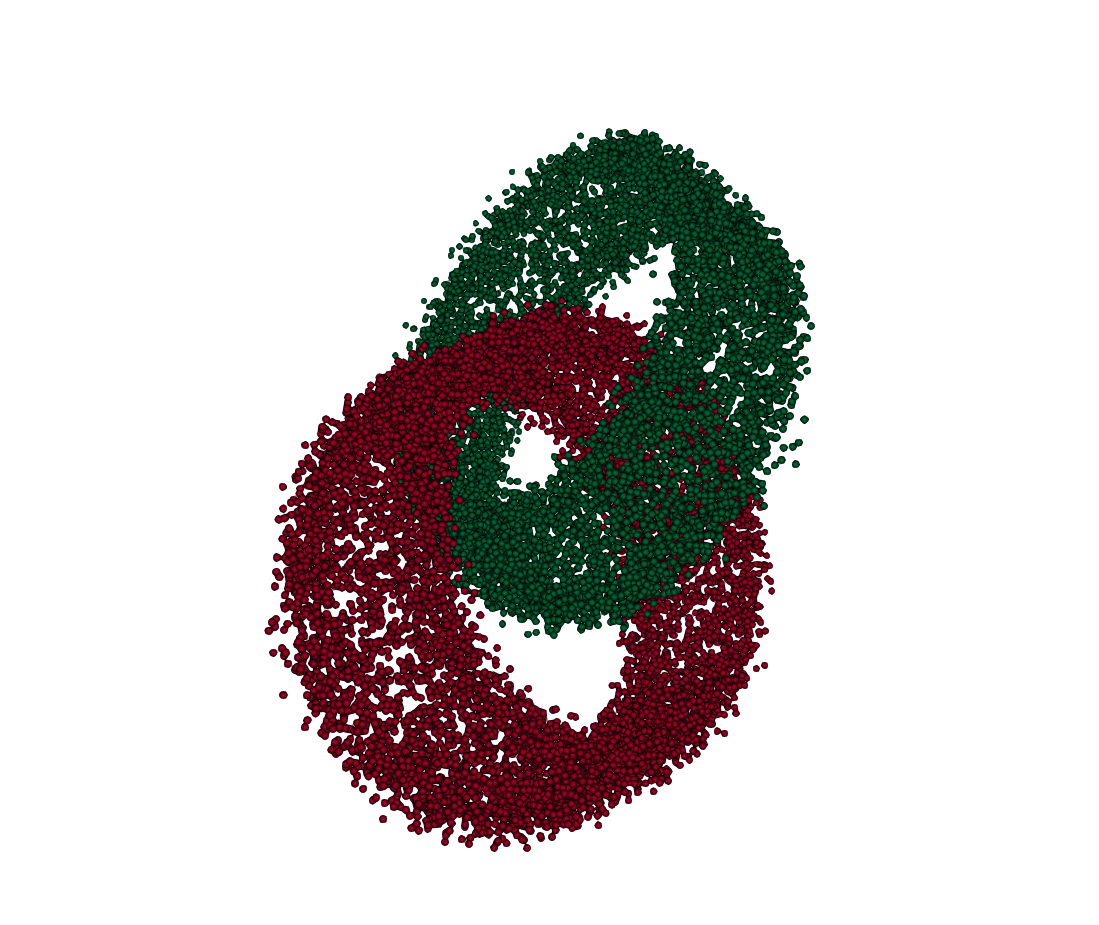}
\includegraphics[trim=3cm 3cm 3cm 3cm, clip, width=0.3\textwidth]{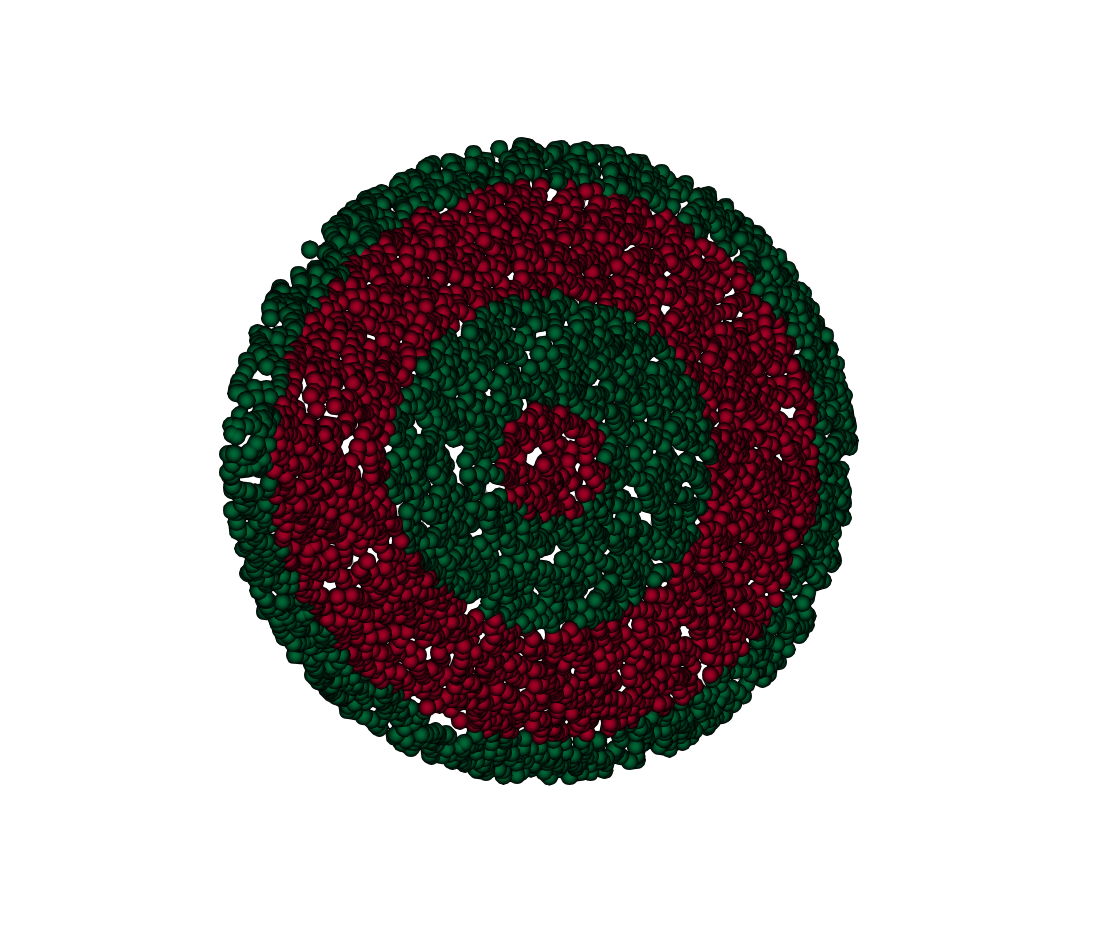}
\includegraphics[trim=4cm 4cm 4cm 4cm, clip, width=0.3\textwidth]{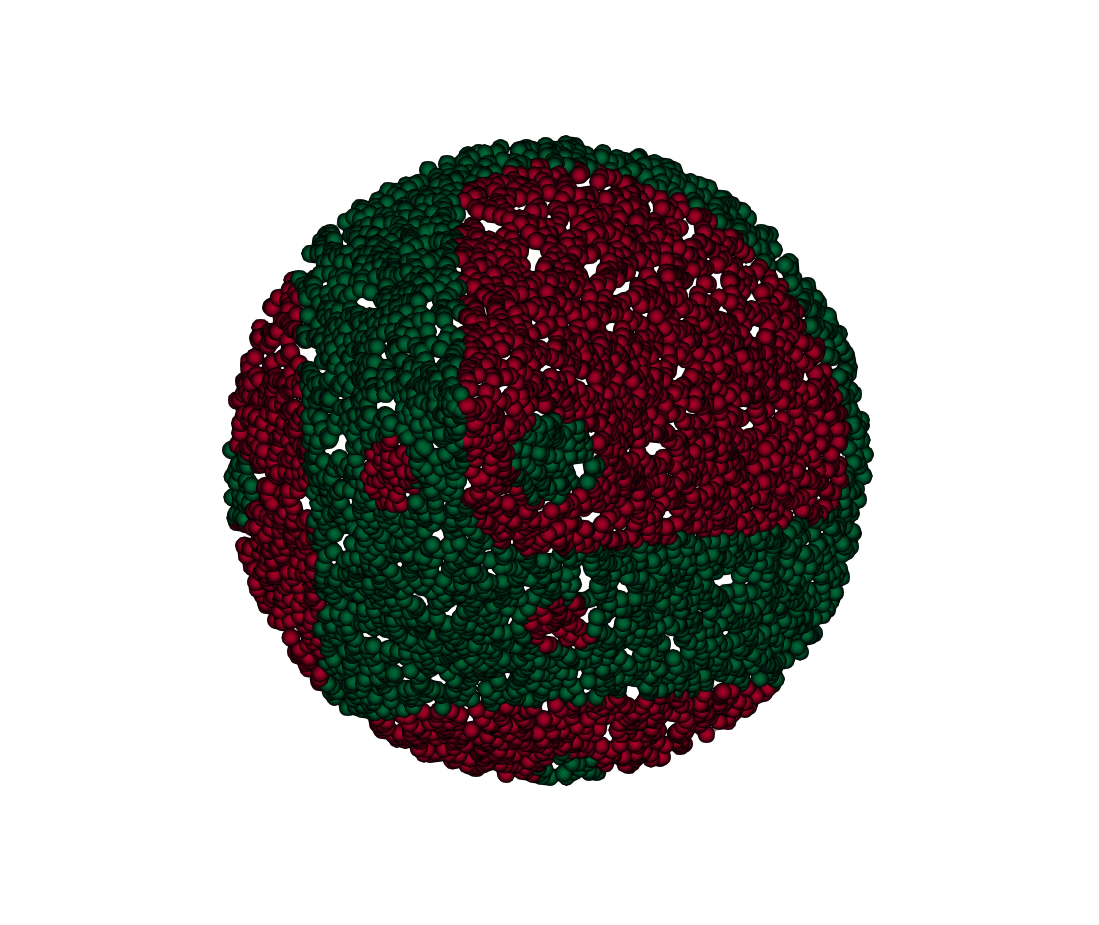}
\caption{\label{fig:rbf} left: ``Tori'' data ($d=3$). Center: Spherical layers ($d=2$). Right: ``RBF'' data ($d=2$).}
\end{figure}

In our first set of experiments, we let $D_i = R(\mathbb T_i \times \mR^{d-3})$ where $\mathbb T_1$, $\mathbb T_2$ are non-intersecting tori in $\mR^3$ and $R$ is a random $d$-dimensional rotation. The tori are positioned as illustrated in the first panel of figure \ref{fig:rbf}, so that, even though they have an empty intersection, they are not linearly separable. The distribution of training and test data is (before rotation) the product of a uniform distribution on the torus and of a standard Gaussian  in the $d-3$ remaining dimensions. 
%\begin{figure}
%\includegraphics[width=0.5\textwidth]{Figures/donuts.png}
%\caption{\label{fig:tori}Tori}
%\end{figure}

\begin{table}[h]
\centering
\begin{tabular}{lcccccccc}%
&Log. reg.&lin. SVM&SVM&RF&kNN&MLP1&MLP2&MLP5\\%
\toprule%
\multicolumn{9}{r}{Tori, $d=3$, 100 samples per class}\\%
\cmidrule{2-9}
Original Data&0.341&0.341&0.000&0.007&0.000&0.004&0.000&0.000\\%
Transformed Data&0.000&0.000&0.000&0.007&0.000&0.000&0.000&0.000\\%
\hlineadd%
\multicolumn{9}{r}{Tori, $d=10$, 100 samples per class}\\%
\cmidrule{2-9}
Original Data&0.312&0.317&0.280&0.311&0.309&0.159&0.170&0.233\\%
Transformed Data&0.159&0.163&0.162&0.176&0.153&0.155&0.153&0.163\\%
\hlineadd%
\multicolumn{9}{r}{Tori, $d=10$, 250 samples per class}\\%
\cmidrule{2-9}
Original Data&0.325&0.326&0.207&0.285&0.257&0.073&0.093&0.109\\%
Transformed Data&0.023&0.024&0.017&0.021&0.012&0.030&0.026&0.025\\%
\hlineadd%
\multicolumn{9}{r}{Tori, $d=20$, 100 samples per class}\\%
\cmidrule{2-9}
Original Data&0.320&0.317&0.324&0.376&0.369&0.325&0.325&0.353\\%
Transformed Data&0.321&0.329&0.322&0.347&0.330&0.320&0.319&0.323\\%
\hlineadd
\multicolumn{9}{r}{Tori, $d=20$, 250 samples per class}\\%
\cmidrule{2-9}
Original Data&0.320&0.323&0.312&0.339&0.367&0.204&0.284&0.280\\%
Transformed Data&0.249&0.255&0.249&0.277&0.251&0.247&0.247&0.250\\%
\hlineadd%
\multicolumn{9}{r}{Tori, $d=20$, 500 samples per class}\\%
\cmidrule{2-9}
Original Data&0.316&0.315&0.267&0.305&0.355&0.117&0.130&0.191\\%
Transformed Data&0.163&0.166&0.158&0.173&0.167&0.160&0.163&0.164\\%
\bottomrule%
\end{tabular}%

\caption{\label{tab:tori} Comparative performance of classifiers on ``Tori'' data}
\end{table}

\begin{figure}[h]
\centering
\includegraphics[trim=1cm 3cm 1cm 3cm, clip, width=0.24\textwidth]{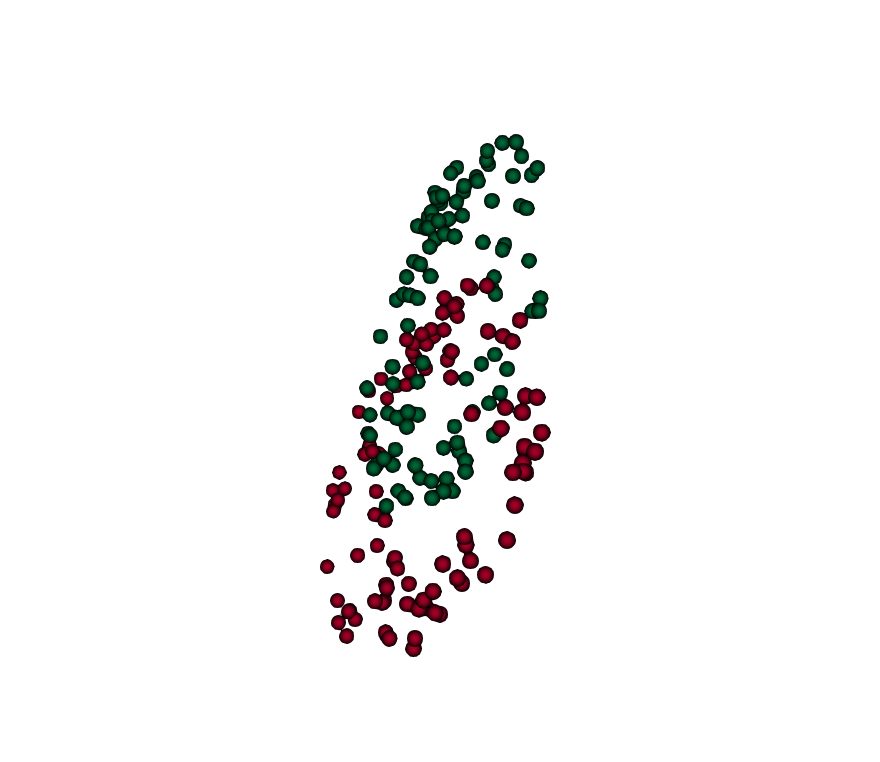}
\includegraphics[trim=1cm 3cm 1cm 3cm, clip, width=0.24\textwidth]{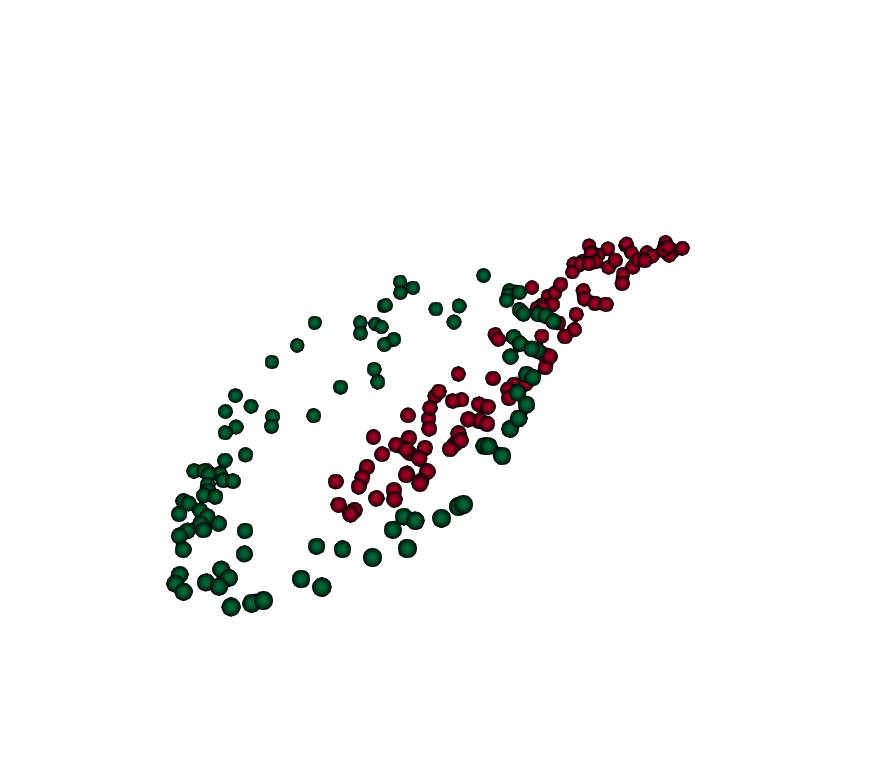}
\includegraphics[trim=1cm 3cm 1cm 3cm, clip, width=0.24\textwidth]{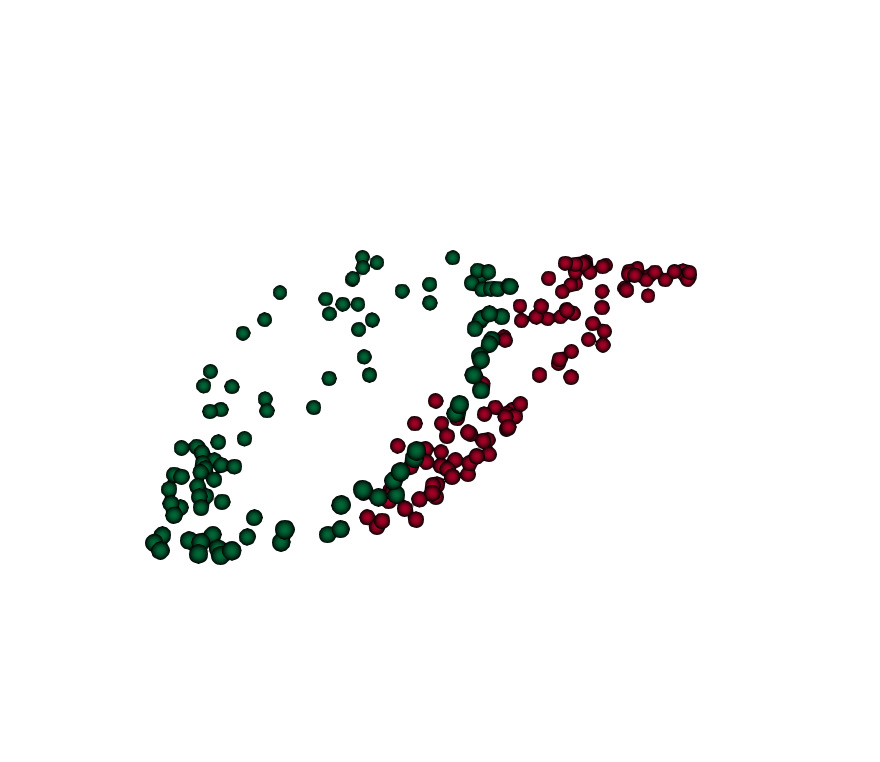}
\includegraphics[trim=1cm 3cm 1cm 3cm, clip, width=0.24\textwidth]{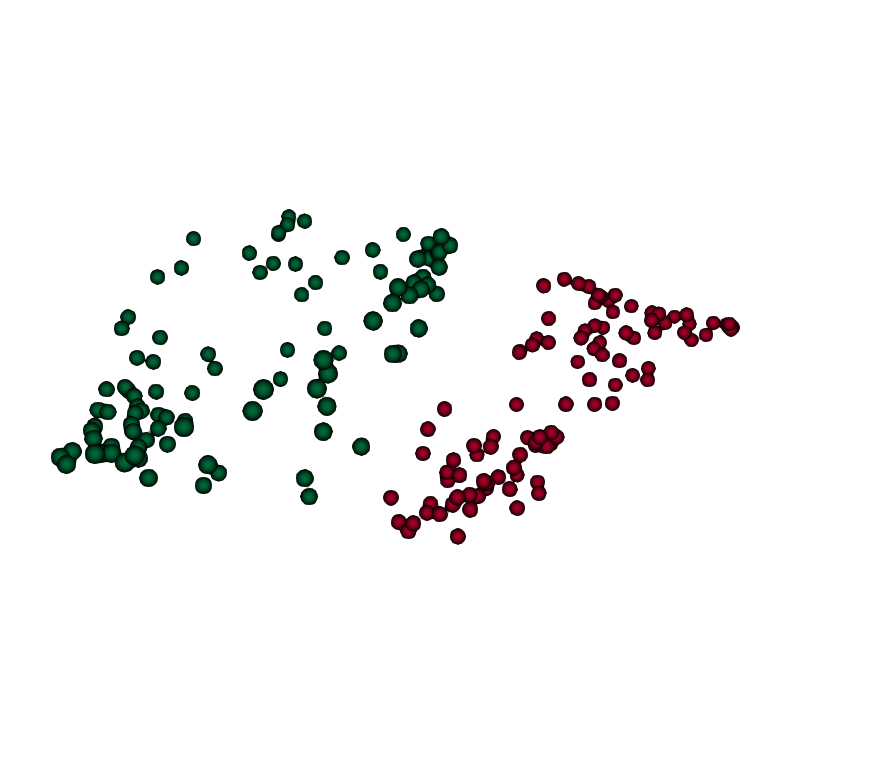}\\
\includegraphics[trim=1cm 3cm 1cm 3cm, clip, width=0.24\textwidth]{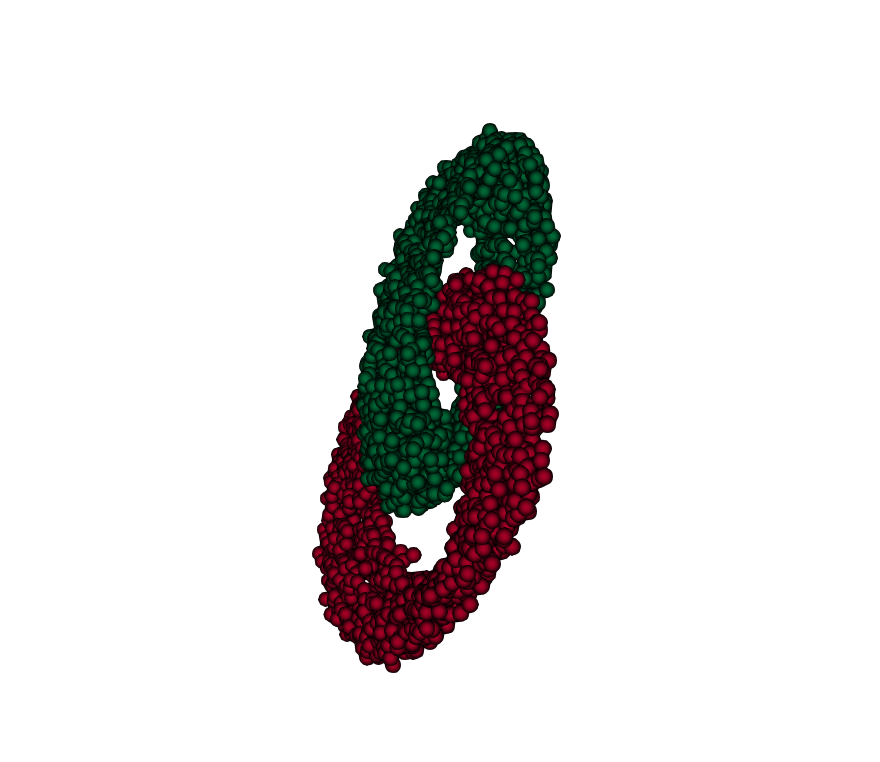}
\includegraphics[trim=1cm 3cm 1cm 3cm, clip, width=0.24\textwidth]{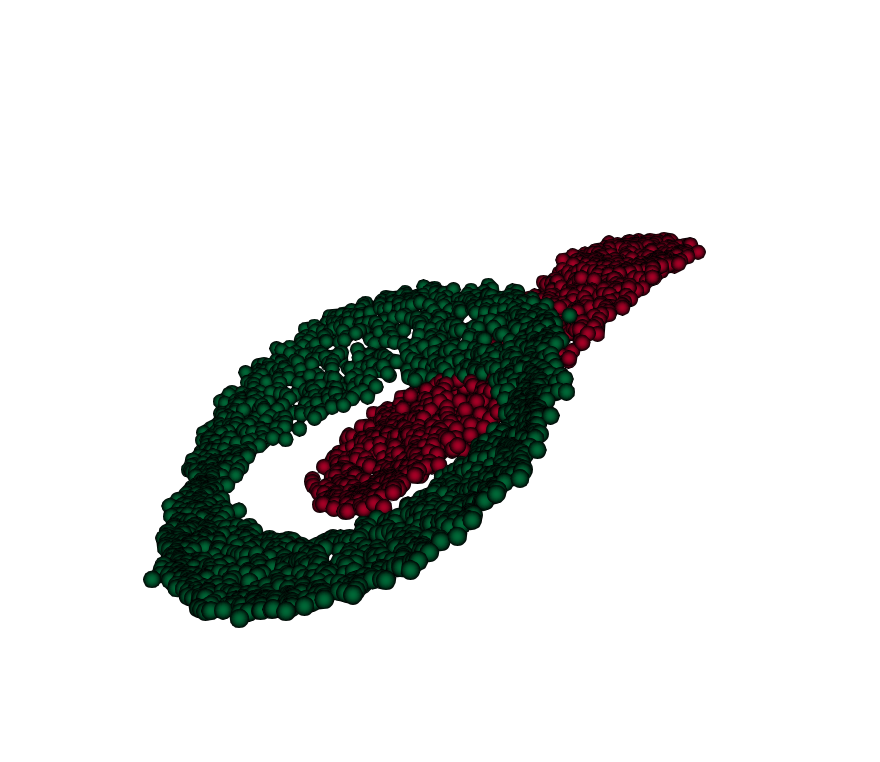}
\includegraphics[trim=1cm 3cm 1cm 3cm, clip, width=0.24\textwidth]{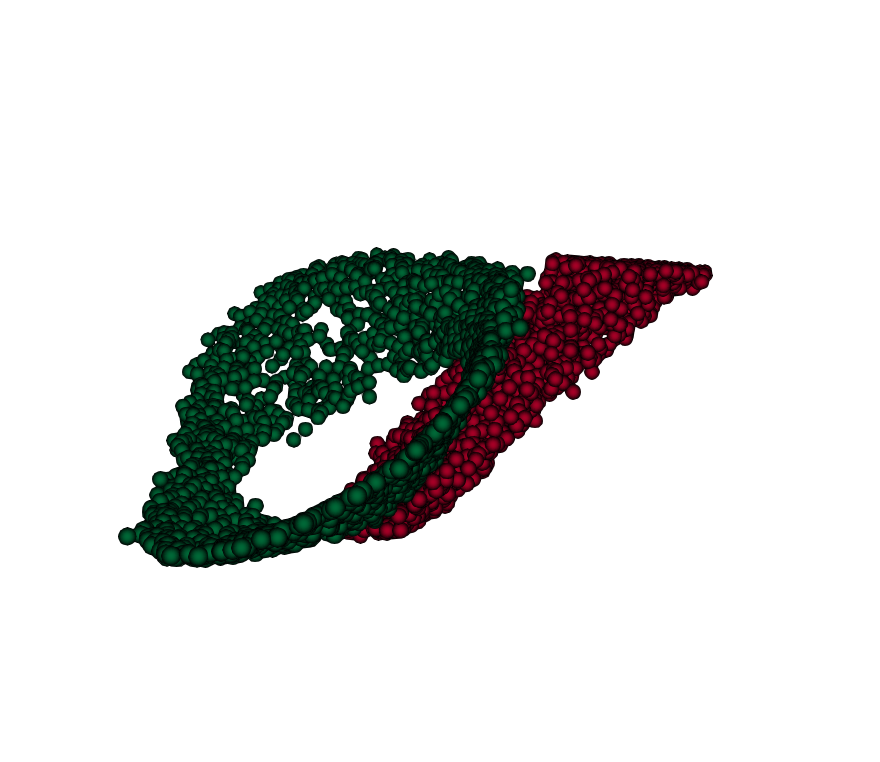}
\includegraphics[trim=1cm 3cm 1cm 3cm, clip, width=0.24\textwidth]{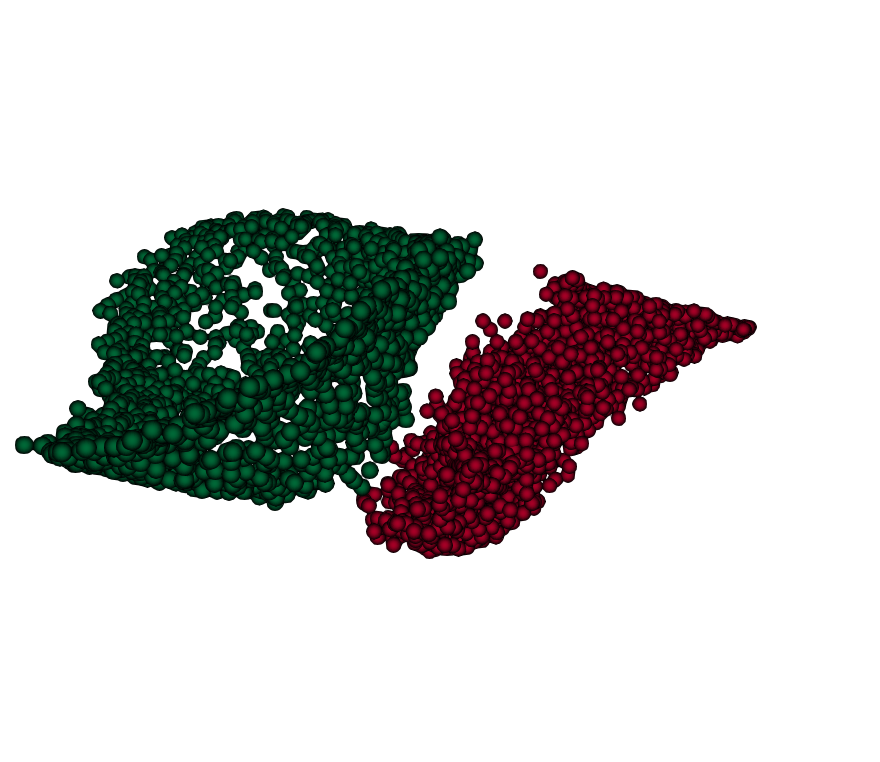}
\caption{\label{fig:tori.diffeo} Visualization of the diffeomorphic flow applied to the 3D tori dataset. Left: training data;  Right: test data.  Top to bottom: $t=0$, $t=0.3$, $t=0.7$, $t=1$. The data is visualized in a frame formed by the discriminant direction followed by the two principal components in the perpendicular space.}
\end{figure}

Classification results for this dataset are summarized in Table \ref{tab:tori}. Here, we let the number of noise dimensions vary from 0 to 7 and 17 (so that the total number of dimensions is 3, 10 and 20) and the number of training samples are 200, 500 and  1,000. The problem becomes very challenging for most classifiers when the number of noisy dimensions is large, in which case a multi-layer perceptron with one hidden layer seems to perform best.  All other classifiers see their performance improved after  diffeomorphic transformation of the data. One can also notice that, after transformation, all classifiers perform approximately at the same level.  Figure \ref{fig:tori.diffeo} illustrates how the data is transformed by the diffeomorphism and is typical of the other results. 

We also point out that all classifiers except RF are invariant by rotation of the data, so that making a random rotation when generating it does not change their performance. The estimation of the diffeomorphism using a radial kernel is also rotation invariant. The RF classifier, which is based on comparison along coordinate axes, is highly affected, however. Without a random rotation, it  performs extremely well, with an error rate of only 0.05 with 17 noisy dimensions and 250 examples per class. 
%Interestingly, if one uses a convolutional kernel to train the diffeomorphism (which exploits a similar bias in the organization of the data), the error rate of linear classifiers after transformation also drop to around 0.003. 

\subsection{Spherical Layers} 
We here deduce the class from the sign of $\cos(9 |X|)$ where $X$ follows a uniform distribution on the $d$-dimensional unit sphere, and we  provide results with $d=3$ (a representation of the data set in 2D is provided in the second panel of Figure \ref{fig:rbf}). We see that linear classifiers cannot do better than chance (this is by design), but that after the estimated diffeomorphic transformation is applied, all classifiers outperform, with a large margin for small datasets, the best non-linear classifiers trained on the original data. 
\begin{table}[h]
\centering
\begin{tabular}{lcccccccc}%
&Log. reg.&lin. SVM&SVM&RF&kNN&MLP1&MLP2&MLP5\\%
\toprule%
\multicolumn{9}{r}{Spherical layers, $d=3$, 100 samples per class}\\%
\cmidrule{2-9} Original Data&0.507&0.507&0.357&0.353&0.408&0.478&0.488&0.481\\%
Transformed Data&0.246&0.249&0.329&0.279&0.355&0.246&0.245&0.253\\%
\hlineadd%
\multicolumn{9}{r}{Spherical layers, $d=3$, 250 samples per class}\\%
\cmidrule{2-9} Original Data&0.487&0.487&0.233&0.231&0.279&0.394&0.264&0.113\\%
Transformed Data&0.119&0.118&0.185&0.129&0.199&0.149&0.143&0.138\\%
\hlineadd%
\multicolumn{9}{r}{Spherical layers, $d=3$, 500 samples per class}\\%
\cmidrule{2-9} Original Data&0.506&0.506&0.185&0.194&0.229&0.299&0.113&0.091\\%
Transformed Data&0.089&0.092&0.165&0.087&0.197&0.092&0.103&0.096\\%
\bottomrule
\end{tabular}%

%\begin{tabular}{ccccccc}
%&Log. Reg. & Lin. SVM & SVM & RF & $k$NN & MLP\\
%\toprule
%&\multicolumn{6}{c}{3  Dimensions, 200  Training samples}\\
%\cmidrule{2-7}
%\cmidrule{2-9} Original Data & 0.498 & 0.498 & 0.439 & 0.394 & 0.390&0.351\\
%\cmidrule{1-7}
%Transformed Data&  0.276 & 0.276 & 0.276 & 0.278 & 0.277& 0.277\\
%\midrule
%&\multicolumn{6}{c}{3  Dimensions, 400  Training samples}\\
%\cmidrule{2-7}
%\cmidrule{2-9} Original Data & 0.482 & 0.483 & 0.364 & 0.288 & 0.307&0.298\\
%\cmidrule{1-7}
%Transformed Data&  0.205 & 0.205 & 0.206 & 0.206 & 0.206&0.206\\
%\midrule
%&\multicolumn{6}{c}{3  Dimensions, 1000  Training samples}\\
%\cmidrule{2-7}
%\cmidrule{2-9} Original Data & 0.501 & 0.501 & 0.289 & 0.203 & 0.228&0.434\\
%\cmidrule{1-7}
%Transformed Data&  0.131 & 0.132 & 0.132 & 0.133 & 0.132&0.133\\
%\midrule
%&\multicolumn{6}{c}{3  Dimensions, 2000  Training samples}\\
%\cmidrule{2-7}
%\cmidrule{2-9} Original Data & 0.477 & 0.477 & 0.246 & 0.196 & 0.189&0.279\\
%\cmidrule{1-7}
%Transformed Data&  0.108 & 0.108 & 0.109 & 0.110 & 0.107&0.107\\
%\bottomrule
%\\[-.1cm]
%\end{tabular}
\caption{\label{tab:dolls} Comparative performance of classifiers on 3D spherical layers.}
\end{table}

\subsection{RBF datasets}
The next dataset generates classes using sums of radial basis functions. More precisely, we let $\rho(z) =\exp(-(z/\alpha)^2)$ with $\alpha = 0.1$ and generate classes according to the sign of the function
\[
\sin\left(\sum_{j=1}^L \rho(|X-c_j|) a_j \right)- \mu\,.
\]
In this expression,  $X$ follows a uniform distribution over the $d$-dimensional unit sphere and $\mu$ is estimated so that both positive and negative classes are balanced. The centers, $c_1, \ldots, c_L$ are chosen as $c_j = (3j/L) e_{(j\, \mathrm {mod} \, d)+1}$ where $j \, \mathrm{mod}\, d$ is the remainder of the division of $j$ by $d$ and $e_1, \ldots, e_d$ is the canonical basis of $\mR^d$. The coefficients are $a_j = \cos(6\pi j/L)$, and we took $L=100$.  The third panel of Figure \ref{fig:rbf} depicts the resulting regions of the unit disc in the case $d=2$.

Table \ref{tab:rbf} provides classification performances for various combinations of dimension and sample size. Excepted when $d=5$ and only 100 samples per class are observed, in which case linear classifiers provide the best rates, the best classification is obtained after diffeomorphic transformation. 

\begin{table}[h]
\centering
\begin{tabular}{lcccccccc}%
&Log. reg.&lin. SVM&SVM&RF&kNN&MLP1&MLP2&MLP5\\%
\toprule
\multicolumn{9}{r}{RBF data, $d=2$, 100 samples per class}\\%
\cmidrule{2-9} Original Data&0.381&0.381&0.158&0.126&0.208&0.183&0.193&0.163\\%
Transformed Data&0.142&0.139&0.149&0.153&0.186&0.140&0.136&0.141\\%
\hlineadd%

\multicolumn{9}{r}{RBF data, $d=2$, 250 samples per class}\\%
\cmidrule{2-9} Original Data&0.358&0.359&0.109&0.150&0.136&0.176&0.158&0.127\\%
Transformed Data&0.101&0.102&0.106&0.102&0.130&0.126&0.122&0.119\\%
\hlineadd%

\multicolumn{9}{r}{RBF data, $d=2$, 500 samples per class}\\%
\cmidrule{2-9} Original Data&0.377&0.378&0.089&0.103&0.097&0.119&0.101&0.081\\%
Transformed Data&0.058&0.055&0.085&0.055&0.090&0.070&0.054&0.142\\%
\hlineadd%

\multicolumn{9}{r}{RBF data, $d=3$, 100 samples per class}\\%
\cmidrule{2-9} Original Data&0.307&0.307&0.209&0.207&0.221&0.218&0.233&0.250\\%
Transformed Data&0.171&0.175&0.202&0.168&0.195&0.162&0.162&0.154\\%
\hlineadd%

\multicolumn{9}{r}{RBF data, $d=3$, 250 samples per class}\\%
\cmidrule{2-9} Original Data&0.310&0.308&0.139&0.182&0.148&0.151&0.151&0.103\\%
Transformed Data&0.083&0.077&0.133&0.082&0.122&0.086&0.083&0.089\\%
\hlineadd%

\multicolumn{9}{r}{RBF data, $d=3$, 500 samples per class}\\%
\cmidrule{2-9} Original Data&0.337&0.335&0.129&0.112&0.114&0.144&0.093&0.094\\%
Transformed Data&0.071&0.070&0.118&0.063&0.102&0.080&0.068&0.078\\%
\hlineadd%

\multicolumn{9}{r}{RBF data, $d=5$, 100 samples per class}\\%
\cmidrule{2-9} Original Data&0.216&0.218&0.221&0.256&0.254&0.280&0.259&0.233\\%
Transformed Data&0.221&0.223&0.221&0.222&0.219&0.221&0.223&0.214\\%
\hlineadd%

\multicolumn{9}{r}{RBF data, $d=5$, 250 samples per class}\\%
\cmidrule{2-9} Original Data&0.233&0.233&0.197&0.226&0.213&0.250&0.273&0.240\\%
Transformed Data&0.199&0.204&0.185&0.215&0.196&0.207&0.209&0.205\\%
\hlineadd%

\multicolumn{9}{r}{RBF data, $d=5$, 500 samples per class}\\%
\cmidrule{2-9} Original Data&0.231&0.230&0.167&0.219&0.184&0.206&0.189&0.158\\%
Transformed Data&0.128&0.129&0.133&0.127&0.142&0.141&0.140&0.143\\%
\bottomrule

\end{tabular}%

%\begin{tabular}{ccccccc}
%&Log. Reg. & Lin. SVM & SVM & RF & $k$NN&MLP\\
%\toprule
%&\multicolumn{6}{c}{10  Dimensions, 200  Training samples}\\
%\cmidrule{2-7}
%\cmidrule{2-9} Original Data & 0.479 & 0.478 & 0.119 & 0.284 & 0.425& 0.276\\
%\cmidrule{1-7}
%Transformed Data&  0.161 & 0.160 & 0.161 & 0.156 & 0.159&0.164\\
%\midrule
%&\multicolumn{6}{c}{10  Dimensions, 400  Training samples}\\
%\cmidrule{2-7}
%\cmidrule{2-9} Original Data & 0.494 & 0.493 & 0.090 & 0.253 & 0.417&0.220\\
%\cmidrule{1-7}
%Transformed Data&  0.094 & 0.093 & 0.094 & 0.097 & 0.093&0.092\\
%\midrule
%&\multicolumn{6}{c}{10  Dimensions, 1000  Training samples}\\
%\cmidrule{2-7}
%\cmidrule{2-9} Original Data & 0.482 & 0.482 & 0.059 & 0.214 & 0.394 &0.148\\
%\cmidrule{1-7}
%Transformed Data&  0.055 & 0.055 & 0.056 & 0.059 & 0.055&0.057\\
%\midrule
%&\multicolumn{6}{c}{10  Dimensions, 2000  Training samples}\\
%\cmidrule{2-7}
%\cmidrule{2-9} Original Data & 0.469 & 0.470 & 0.040 & 0.184 & 0.346&0.071\\
%\cmidrule{1-7}
%Transformed Data&  0.044 & 0.043 & 0.043 & 0.043 & 0.042&0.042\\
%\bottomrule
%\\[-.1cm]
%\end{tabular}
\caption{\label{tab:rbf} Comparative performance of classifiers on ``RBF'' data}
\end{table}

%(Nonlinear) SVM performs best on this  dataset, which is not too surprising, because it is essentially tailored for this kind of problem. Linear classifiers are barely better than chance before diffeomorphic transformation. If one excepts the smallest dataset (200 training samples) in which a 5\% gap can be observed, the performance of the linear classifiers after transformations are quite similar to those of non-linear  SVM before  transformation.

\subsection{Mixture of Gaussians}

\begin{figure}[h]
\centering
\includegraphics[trim= 4cm 6cm 6cm 6cm, clip,width=0.19\textwidth]{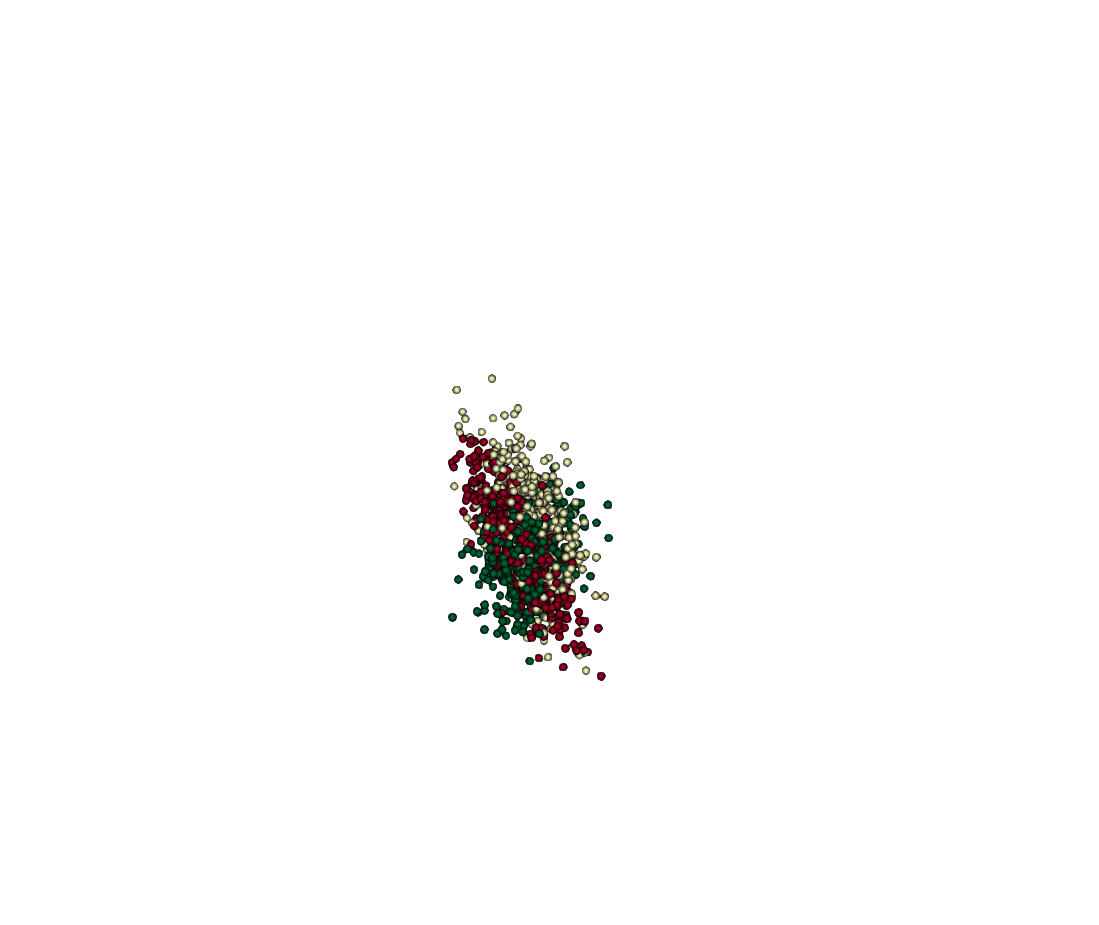}
\includegraphics[trim= 4cm 6cm 6cm 6cm, clip,width=0.19\textwidth]{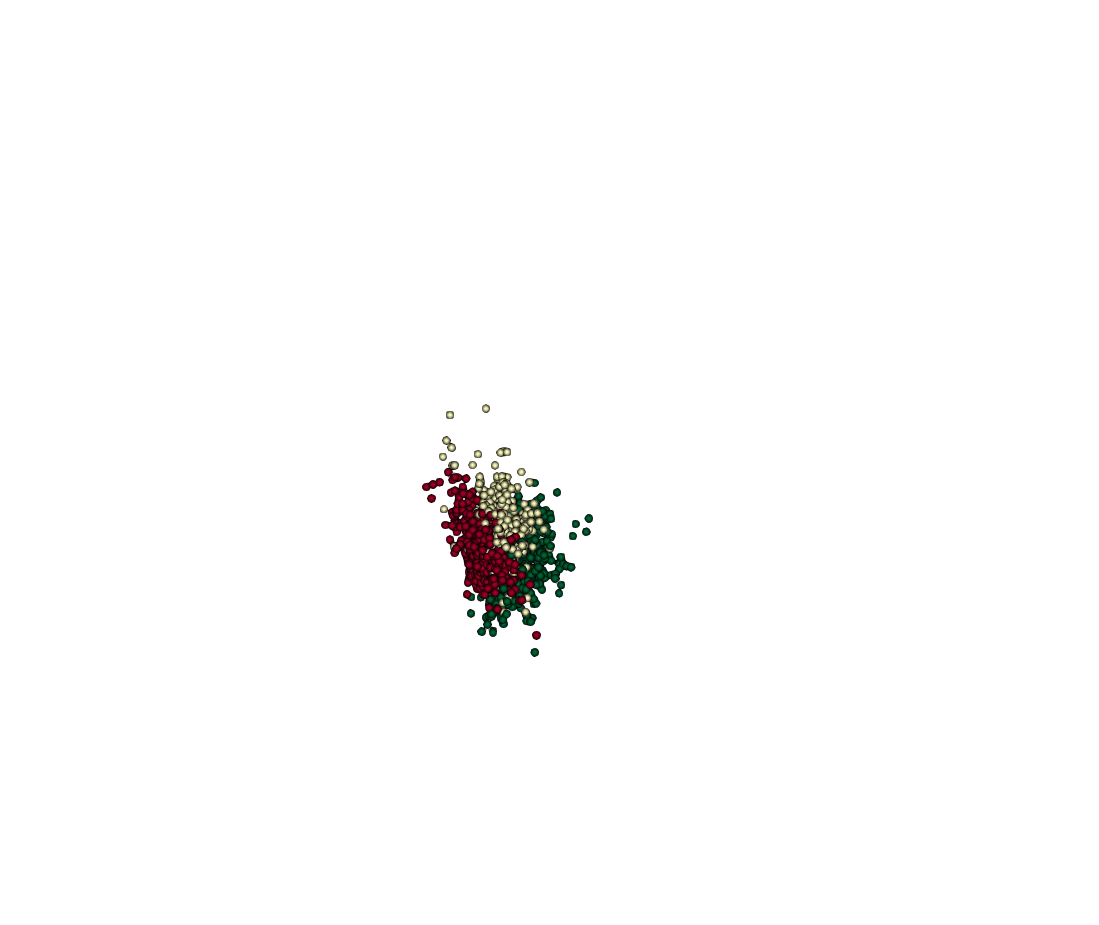}
\includegraphics[trim= 4cm 6cm 6cm 6cm, clip,width=0.19\textwidth]{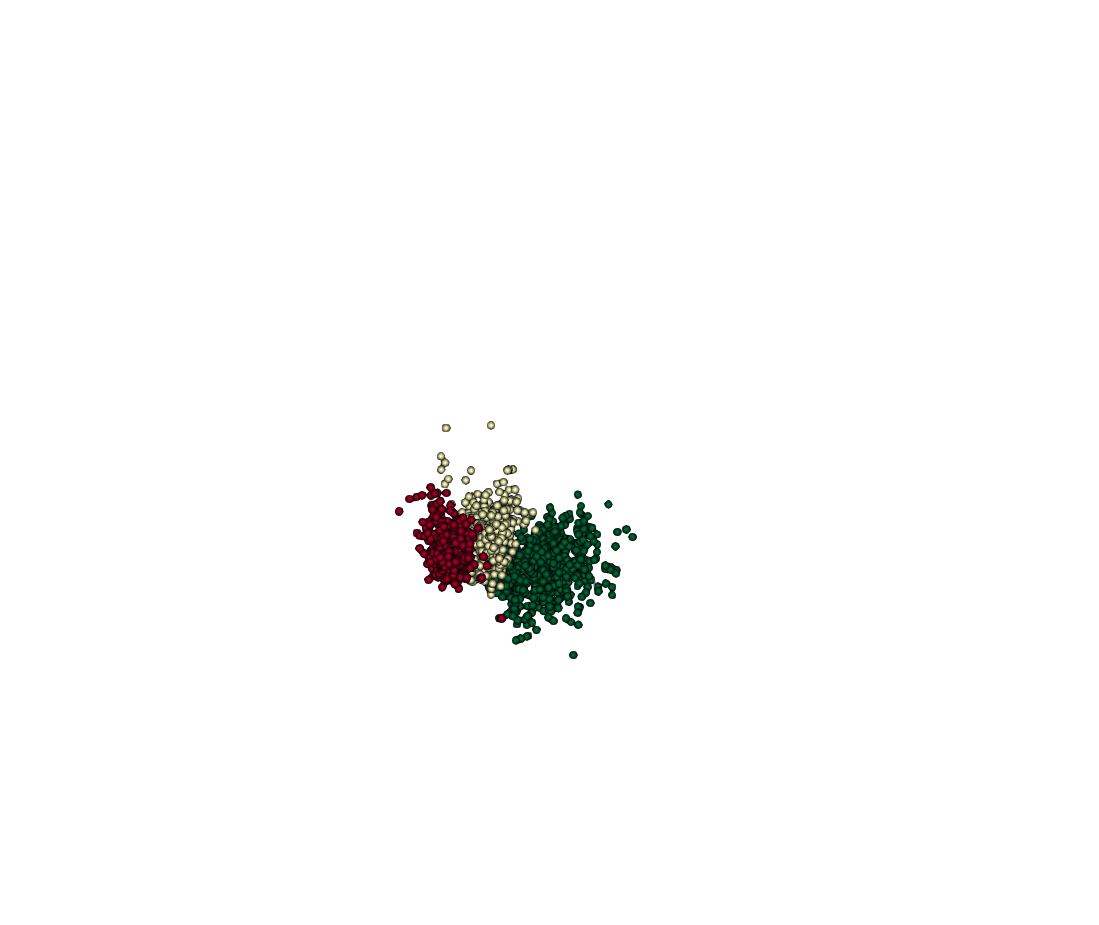}
\includegraphics[trim= 4cm 6cm 6cm 6cm, clip,width=0.19\textwidth]{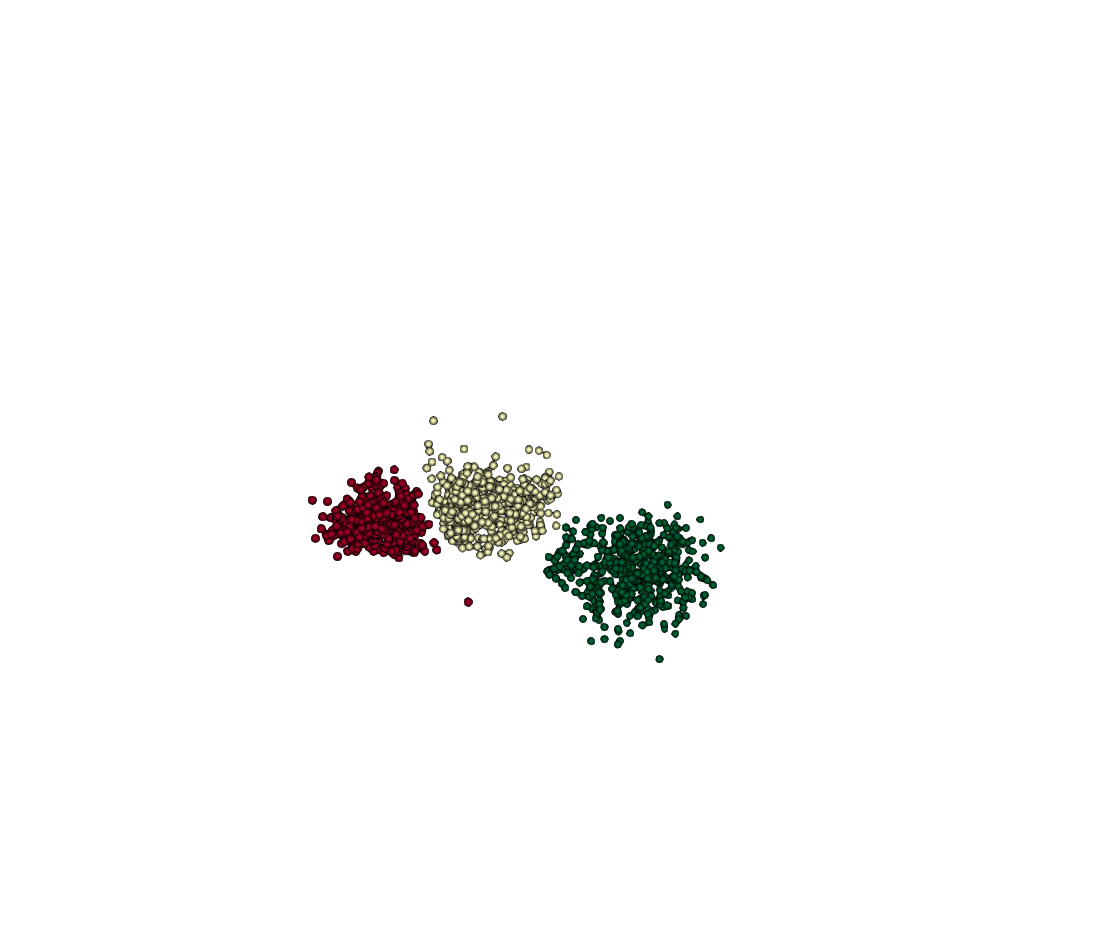}
\includegraphics[trim= 4cm 6cm 6cm 6cm, clip, width=0.19\textwidth]{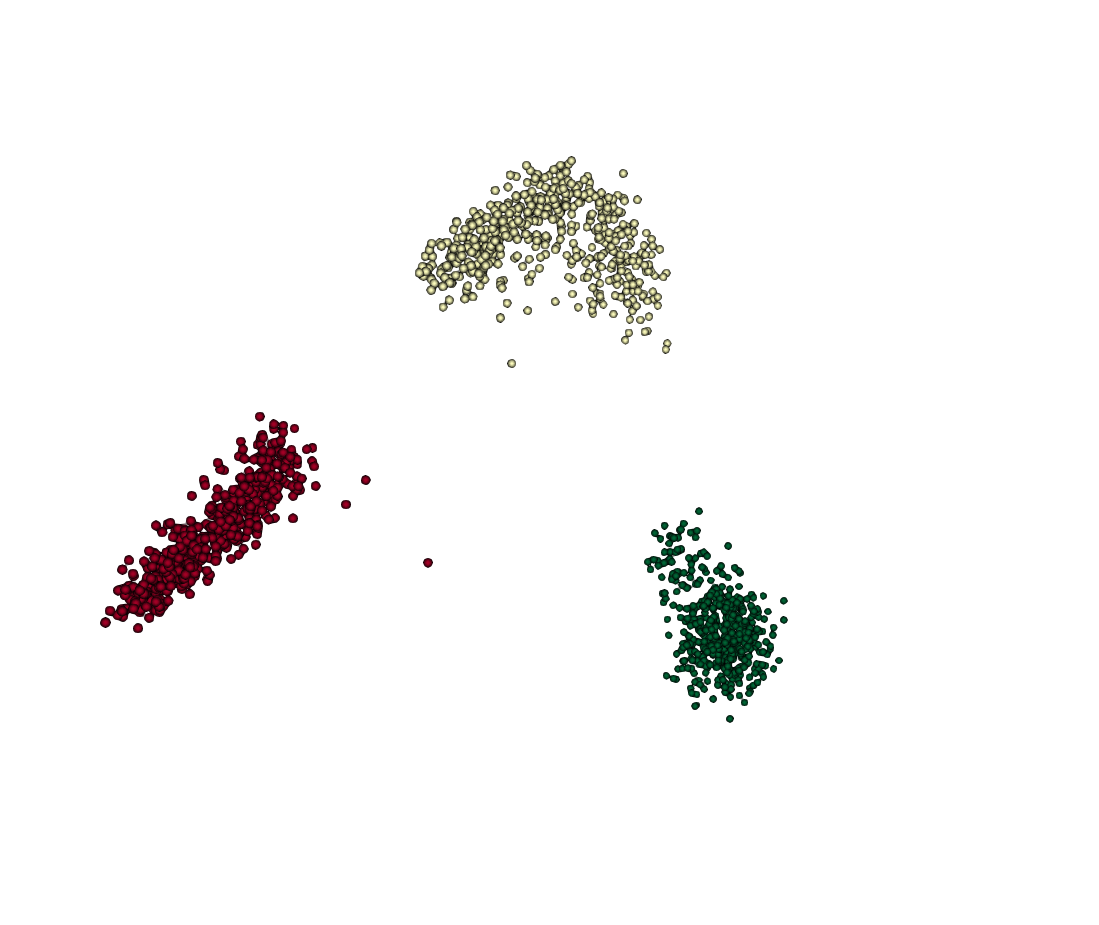}\\
\includegraphics[trim= 4cm 6cm 6cm 6cm, clip,width=0.19\textwidth]{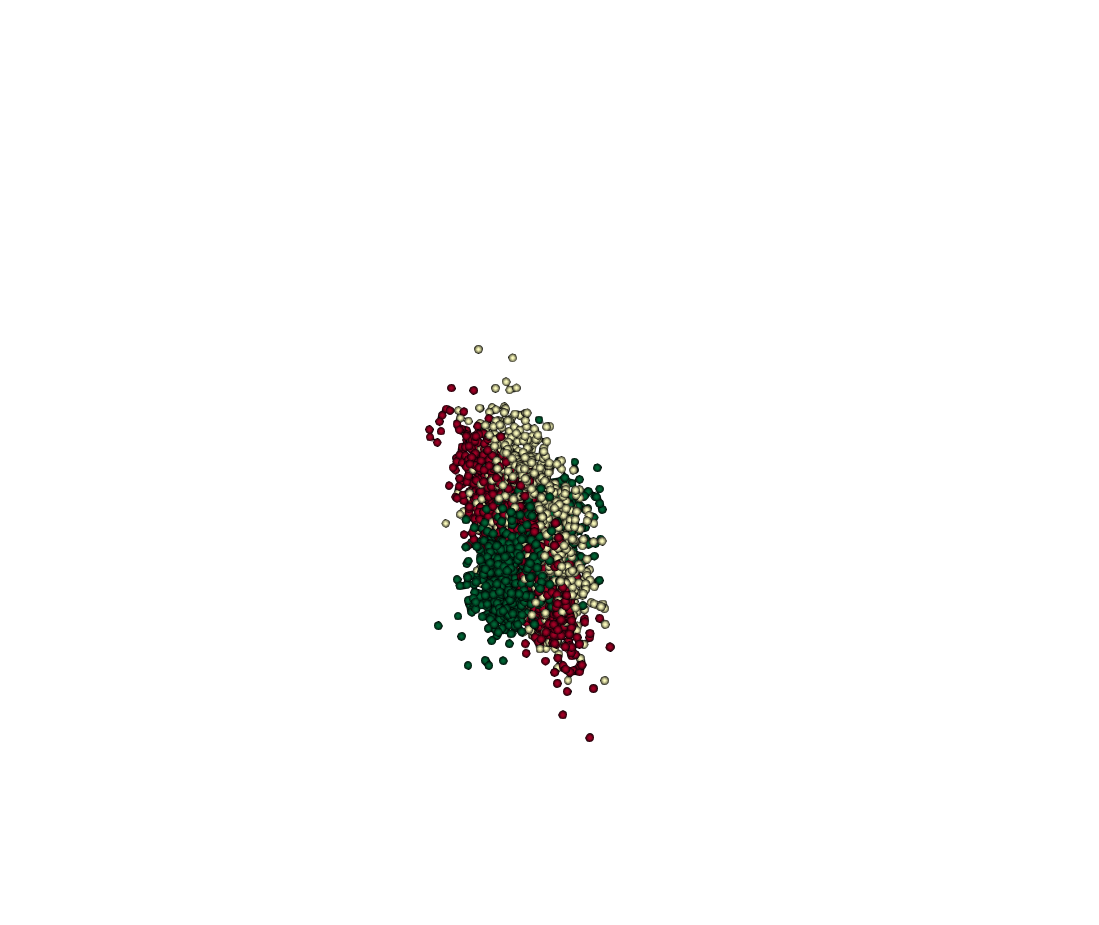}
\includegraphics[trim= 4cm 6cm 6cm 6cm, clip,width=0.19\textwidth]{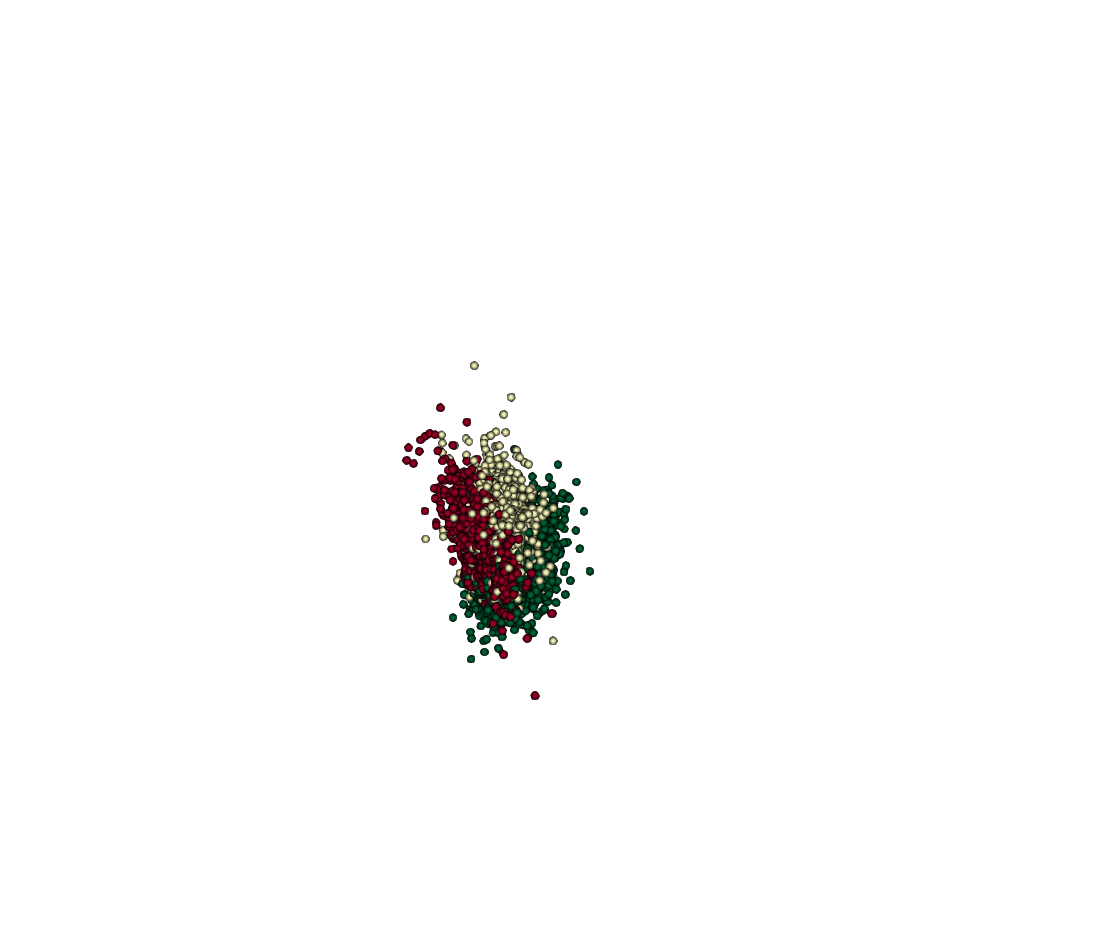}
\includegraphics[trim= 4cm 6cm 6cm 6cm, clip,width=0.19\textwidth]{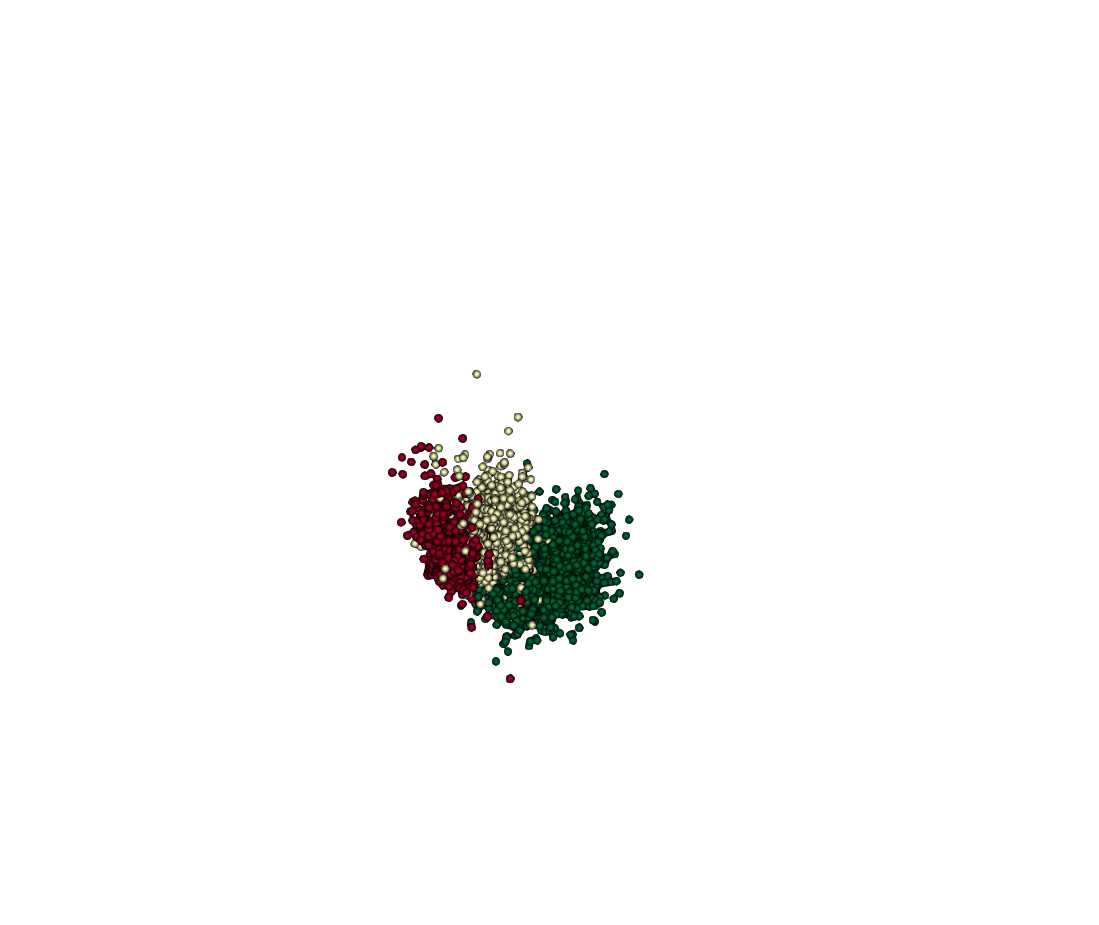}
\includegraphics[trim= 4cm 6cm 6cm 6cm, clip,width=0.19\textwidth]{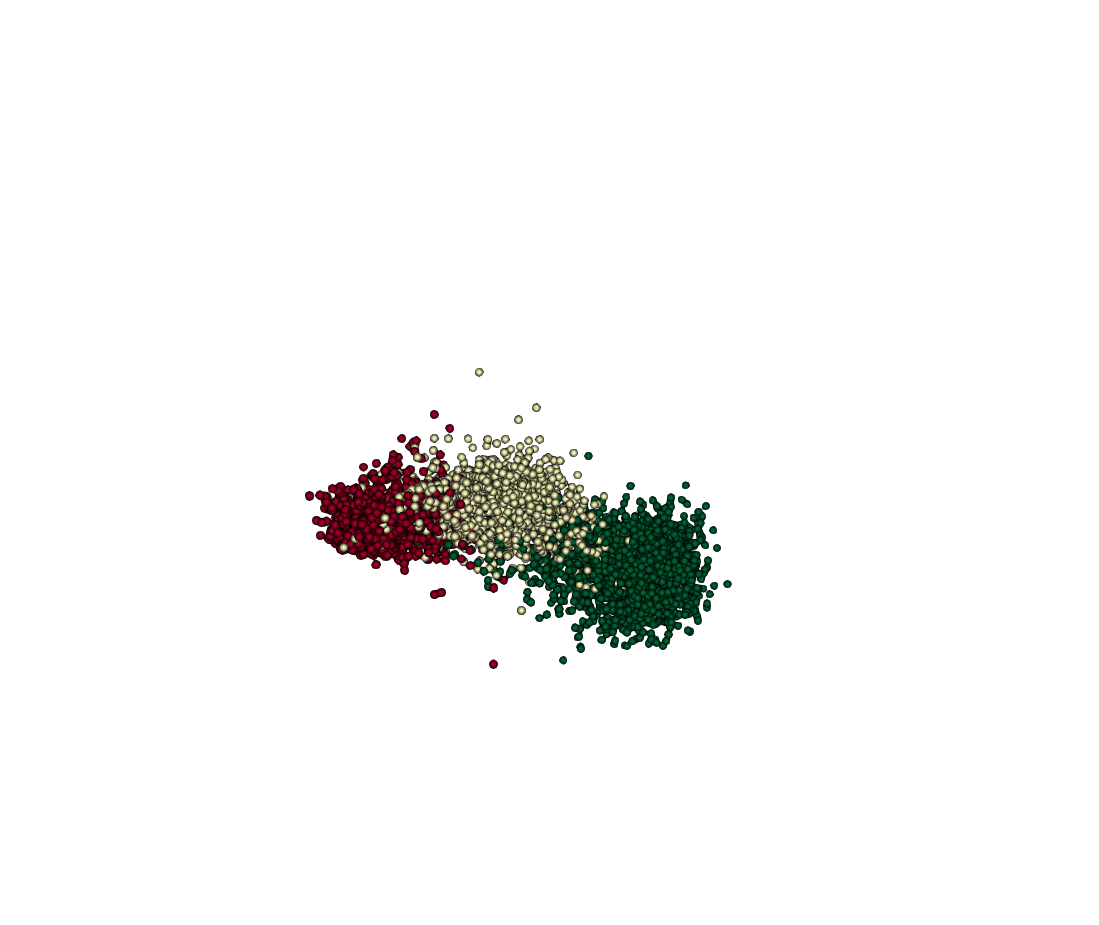}
\includegraphics[trim= 4cm 6cm 6cm 6cm, clip, width=0.19\textwidth]{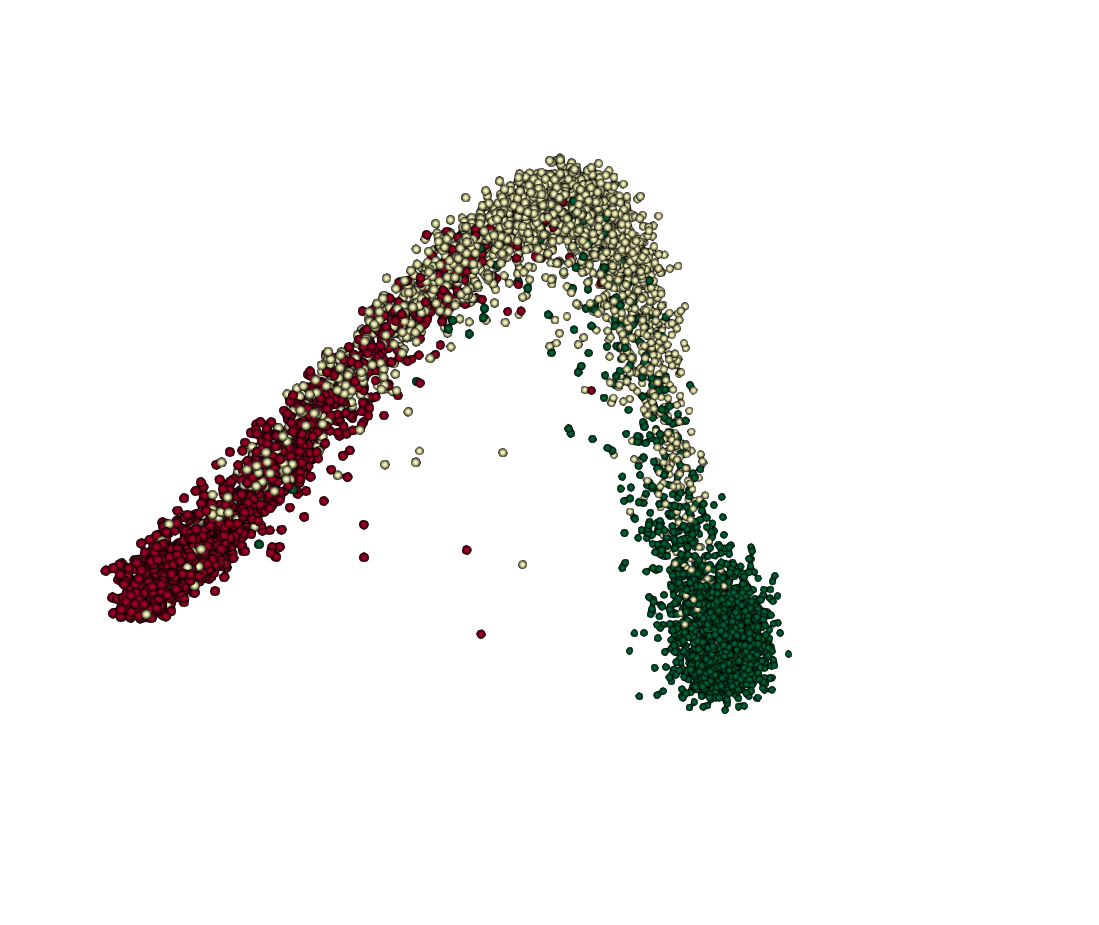}\\
\caption{\label{fig:MoG} Diffeomorphic transformation applied to the Gaussian mixture dataset (evolution visualized at times $t=0.0, 0.3, 0.5, 0.7$ and 1.0. The data is 20 dimensional, and the visualization is made in the plane generated by the two discriminative directions provided by logistic regression estimated on transformed data at time $t=1.$}
\end{figure}

In the next example, we assume that the conditional distribution of $X$ given $Y=y$ is normal with mean $m_y$ and covariance matrix $\Sigma_y$. We used three classes ($y \in \{1,2,3\}$) in dimension 20, with 
\begin{enumerate}[label=$\bullet$]
\item $m_1 = (0, \ldots, 0)^T$, $m_2 = (-1,-1,-1,0, \ldots, 0)^T$ and $m^3 = (-1,1,-1,0. \ldots,0)^T$,
\item $\Sig_1 = 10 \,\Id_{\mR^d}$, $\Sig_2(i,j) = 20 \exp(- |i-j|/20)$ and $\Sig_3(i,j) = 20 \exp(- |i-j|/60)$,
\end{enumerate}
where $\Id_{\mR^d}$ is the $d$-dimensional identity matrix.
Classification results for training sets with 100 and 250 examples per class are described in Table \ref{tab:mog}. While multilayer perceptrons perform best on this data, the performance of diffeomorphic classification is comparable, and improves on all other classifiers. A visualization of the transformation applied to this dataset is provided in Figure \ref{fig:MoG}.

\begin{table}[h]
\centering
 \begin{tabular}{lcccccccc}%
&Log. reg.&lin. SVM&SVM&RF&kNN&MLP1&MLP2&MLP5\\%
\toprule
\multicolumn{9}{r}{Mixture of Gaussians, $d=20$, 100 samples per class}\\%
\cmidrule{2-9} Original Data&0.398&0.407&0.225&0.191&0.495&0.143&0.101&0.099\\%
Transformed Data&0.135&0.153&0.175&0.110&0.236&0.134&0.109&0.108\\%
\hlineadd%
\multicolumn{9}{r}{Mixture of Gaussians, $d=20$, 250 samples per class}\\%
\cmidrule{2-9} Original Data&0.354&0.359&0.172&0.163&0.427&0.073&0.061&0.063\\%
Transformed Data&0.079&0.091&0.145&0.089&0.187&0.077&0.074&0.074\\%
\bottomrule
\end{tabular}%
\caption{\label{tab:mog} Comparative performance of classifiers on Gaussian mixtures.}
\end{table}

\subsection{Curve Segments}

To build this dataset, we start with two scalar functions and assume that the observed data is the restriction of one of them (where this choice determines the class) to a small discrete interval. More precisely, we let $Y\in \{1,2\}$ and given $Y=y$, we let  $X = (\psi_y(t-A) + \varepsilon_t, t\in I)$ where $I = \{0, 1/d, \ldots, (d-1/d)\}$ is a discretization of the unit interval, $(\varepsilon_t, t\in I)$ are independent standard Gaussian variables and $A$ follows a uniform distribution on $[-2, 2]$. For the experiments in Table \ref{tab:curv.seg}, we took $\psi_y(t) = \log\cosh(t/\beta_y)$ with $\beta_1 = 0.02$ and $\beta_2 = 0.021$. This table shows significant improvements of classification rates after application of the diffeomorphism.
\begin{table}[h]
\centering
\begin{tabular}{lcccccccc}%
&Log. reg.&lin. SVM&SVM&RF&kNN&MLP1&MLP2&MLP5\\%
\toprule
\multicolumn{9}{r}{Curve segments, $d=50$, 100 samples per class}\\%
\cmidrule{2-9} Original Data&0.484&0.495&0.400& 0.448&0.466&0.449& 0.444& 0.465\\%
Transformed Data& 0.324&0.324&0.350&0.378&0.438&0.329&0.335&0.332\\
\hlineadd%
\multicolumn{9}{r}{Curve segments, $d=50$, 250 samples per class}\\%
\cmidrule{2-9} Original Data& 0.487 &0.486 &0.266&0.345&0.353&0.493&0.486&0.467\\
Transformed Data&  0.152&0.157&0.147&0.195&0.248&0.155&0.258&0.216\\
\hlineadd%
\multicolumn{9}{r}{Curve segments, $d=50$, 500 samples per class}\\%
\cmidrule{2-9} Original Data&0.495&0.496&0.155&0.294&0.230&0.505&0.484&0.426\\
Transformed Data& 0.093&0.094&0.062&0.070&0.093&0.093&0.104&0.113\\
\bottomrule
\end{tabular}%
\caption{\label{tab:curv.seg} Comparative performance of classifiers on "curve segment" data.}
\end{table}

\subsection{Xor}
Here, we consider a 50-dimensional dataset with two classes. Each sample has all coordinates equal to zero except two of them that are equal to $\pm1$, and the class is 0 if the two nonzero coordinates are equal and 1 otherwise. The position of the two nonzero values is random and each takes the values $\pm1$ with equal probability.

Table \ref{tab:xor} show that diffeomorphic classification performs well on this data with 100 training samples per class. When the number of examples is raised to 250 per class, multi-layer perceptrons with one or two hidden layers perform well as well. The other classifiers perform poorly even on the larger dataset. The classes are obviously not linearly separable, explaining the performances of logistic regression and linear SVMs, and the minimum distance between distinct examples from the same class is the same as that between examples from different classes, namely $\sqrt 2$.

\begin{table}[h]
\centering
\begin{tabular}{lcccccccc}%
&Log. reg.&lin. SVM&SVM&RF&kNN&MLP1&MLP2&MLP5\\%
\toprule
\multicolumn{9}{r}{xor data, $d=50$, 100 samples per class}\\%
\cmidrule{2-9} Original Data&0.505&0.511&0.484&0.500&0.499&0.345&0.428&0.493\\%
Transformed Data&0.066&0.068&0.010&0.024&0.017&0.019&0.016&0.014\\%
\hlineadd%

\multicolumn{9}{r}{xor data, $d=50$, 250 samples per class}\\%
\cmidrule{2-9} Original Data&0.498&0.497&0.404&0.461&0.494&0.005&0.047&0.248\\%
Transformed Data&0.008&0.006&0.000&0.000&0.000&0.000&0.000&0.001\\%
\bottomrule
\end{tabular}%
\caption{\label{tab:xor} Comparative performance of classifiers on xor data.}
\end{table}

\subsection{Segment Lengths}
Our next synthetic example is a simple pattern recognition problem in $d = 100$ dimensions. Samples from class 1 take the form $(\rho_1 U_1, \ldots, \rho_d U_d) $ where $\rho$  is uniformly distributed between $0.75$ and $1.25$, and $U = (U_1, \ldots U_d)$ is a binary vector with exactly $l_y$ consecutive ones (possibly wrapping around $\{1, \ldots, d\}$)and $d-l_y$ zeros, with $l_1=10$ and $l_2=11$.  

An optimal linear separation between classes is achieved (because of shift invariance) by thresholding the sum of the $d$ variables. However, because of the multiplication by $\rho_1, \ldots, \rho_d$, this simple classification rule performs very poorly, while the same rule applied to the binary variables obtained by thresholding individual entries at, say, 0.5 perfectly separates the classes. It is therefore not surprising that multi-layer perceptrons perform very well on this problem and achieve the best classification rates. All other classifiers  perform significantly better when run on the transformed data.

\begin{table}[h]
\centering
\begin{tabular}{lcccccccc}%
&Log. reg.&lin. SVM&SVM&RF&kNN&MLP1&MLP2&MLP5\\%
\toprule
\multicolumn{9}{r}{Segment length data, $d=100$, 100 samples per class}\\%
\cmidrule{2-9} 
Original Data&0.462&0.412&0.387&0.491&0.505&0.337&0.343&0.328\\%
Transformed Data&0.341&0.349&0.354&0.347&0.450&0.390&0.385&0.362\\%
\hlineadd%
\multicolumn{9}{r}{Segment length data, $d=100$, 250 samples per class}\\%
\cmidrule{2-9} 
Original Data&0.432&0.412&0.135&0.483&0.503&0.082&0.079&0.084\\%
Transformed Data&0.100&0.098&0.107&0.098&0.443&0.096&0.113&0.114\\%
\hlineadd%
\multicolumn{9}{r}{Segment length data, $d=100$, 500 samples per class}\\%
\cmidrule{2-9} 
Original Data&0.391&0.391&0.045&0.447&0.194&0.025&0.025&0.025\\%
Transformed Data&0.032&0.030&0.032&0.054&0.165&0.029&0.031&0.036\\%
\bottomrule
\end{tabular}%

%\begin{tabular}{ccccccc}
%&Log. Reg. & Lin. SVM & SVM & RF & $k$NN &MLP\\
%\toprule
%&\multicolumn{6}{c}{100  Dimensions, 200  Training samples}\\
%\cmidrule{2-7}
%\cmidrule{2-9} Original Data & 0.451 & 0.448 & 0.470 & 0.474 & 0.492&0.312\\
%\cmidrule{1-7}
%Transformed Data&  0.312 & 0.312 & 0.312 & 0.324 & 0.312&0.312\\
%\midrule
%&\multicolumn{6}{c}{100  Dimensions, 400  Training samples}\\
%\cmidrule{2-7}
%\cmidrule{2-9} Original Data & 0.439 & 0.443 & 0.477 & 0.471 & 0.502&0.216\\
%\cmidrule{1-7}
%Transformed Data&  0.187 & 0.187 & 0.186 & 0.201 & 0.187&0.186\\
%\midrule
%&\multicolumn{6}{c}{100  Dimensions, 1000  Training samples}\\
%\cmidrule{2-7}
%\cmidrule{2-9} Original Data & 0.426 & 0.433 & 0.469 & 0.462 & 0.321&0.020\\
%\cmidrule{1-7}
%Transformed Data&  0.025 & 0.025 & 0.025 & 0.0367 & 0.025&0.025\\
%\midrule
%&\multicolumn{6}{c}{100  Dimensions, 2000  Training samples}\\
%\cmidrule{2-7}
%\cmidrule{2-9} Original Data & 0.444 & 0.442 & 0.433 & 0.396 & 0.074&0.0\\
%\cmidrule{1-7}
%Transformed Data&  0.004 & 0.004 & 0.004 & 0.009 & 0.004& 0.004\\
%\bottomrule
%\\[-0.1cm]
%\end{tabular}
\caption{\label{tab:segments} Comparative performance of classifiers on segment lengths}
\end{table}

\subsection{Segment Pairs}
This section describes a more challenging version of the former in which each data point consists in two sequences of ones in the unit circle (discretized over 50 points), these two sequences being both of length five in class 1, and of lengths 4 and 6 (in any order) in class 2.
No linear rule can separate the classes and do better than chance, since such a rule would need to be based on summing the variables (by shift invariance), which returns 10 in both classes.  The problem is also challenging for metric-based methods  because each data point has close neighbors in the other class: the nearest non-identical neighbor in the same class is obtained by shifting one of the two segments, and would be at distance $\sqrt2$, which is identical to the distance of the closest element from the other class which is obtained by replacing a point in one segment by 0 and adding a 1 next to the other segment. One way to separate the classes would be to compute moving averages (convolutions) over windows of lengths 6 along the circle, and to threshold the result at 5.5, which can be represented as a one-hidden layer perceptron.
As shown in Table \ref{tab:segment.pairs}, one-hidden-layer perceptrons do perform best on this data, with some margin compared to all others. Diffeomorphic classification does however improve significantly on the classification rates of all other classifiers. 

\begin{table}[h]
\centering
\begin{tabular}{lcccccccc}%
&Log. reg.&lin. SVM&SVM&RF&kNN&MLP1&MLP2&MLP5\\%
\toprule
\multicolumn{9}{r}{Segment pair data, $d=50$, 100 samples per class}\\%
\cmidrule{2-9} Original Data&0.477&0.466&0.476&0.475&0.458&0.470&0.470&0.488\\%
Transformed Data&0.461&0.457&0.465&0.461&0.469&0.461&0.462&0.461\\%
\hlineadd%

\multicolumn{9}{r}{Segment pair data, $d=50$, 250 samples per class}\\%
\cmidrule{2-9} Original Data&0.490&0.485&0.447&0.412&0.492&0.343&0.392&0.405\\%
Transformed Data&0.394&0.392&0.412&0.390&0.444&0.398&0.391&0.386\\%
\hlineadd%

\multicolumn{9}{r}{Segment pair data, $d=50$, 500 samples per class}\\%
\cmidrule{2-9} Original Data&0.501&0.502&0.369&0.320&0.504&0.054&0.164&0.284\\%
Transformed Data&0.240&0.238&0.261&0.233&0.378&0.250&0.244&0.253\\%
\hlineadd%

\multicolumn{9}{r}{Segment pair data, $d=50$, 1000 samples per class}\\%
\cmidrule{2-9} Original Data&0.465&0.463&0.233&0.191&0.547&0.008&0.038&0.137\\%
Transformed Data&0.071&0.068&0.099&0.073&0.281&0.085&0.082&0.085\\%
\bottomrule%
\end{tabular}%

%\begin{tabular}{ccccccc}
%&Log. Reg. & Lin. SVM & SVM & RF & $k$NN &MLP\\
%\toprule
%&\multicolumn{6}{c}{100  Dimensions, 200  Training samples}\\
%\cmidrule{2-7}
%\cmidrule{2-9} Original Data & 0.513 & 0.515 & 0.460 & 0.505 & 0.532&0.411\\
%\cmidrule{1-7}
%Transformed Data&  0.421 & 0.421 & 0.421 & 0.422 & 0.421&0.421\\
%\midrule
%&\multicolumn{6}{c}{100  Dimensions, 400  Training samples}\\
%\cmidrule{2-7}
%\cmidrule{2-9} Original Data & 0.465 & 0.467 & 0.498 & 0.497 & 0.488&0.144\\
%\cmidrule{1-7}
%Transformed Data&  0.303 & 0.303 & 0.303 & 0.303 & 0.303&0.302\\
%\midrule
%&\multicolumn{6}{c}{100  Dimensions, 1000  Training samples}\\
%\cmidrule{2-7}
%\cmidrule{2-9} Original Data & 0.543 & 0.549 & 0.450 & 0.499 & 0.403&0.024\\
%\cmidrule{1-7}
%Transformed Data&  0.146 & 0.146 & 0.145 & 0.145 & 0.143&0.145\\
%\midrule
%&\multicolumn{6}{c}{100  Dimensions, 2000  Training samples}\\
%\cmidrule{2-7}
%\cmidrule{2-9} Original Data & 0.514 & 0.512 & 0.442 & 0.510 & 0.283&0.013\\
%\cmidrule{1-7}
%Transformed Data&  0.069 & 0.069 & 0.069 & 0.070 & 0.068& 0.068\\
%\bottomrule
%\\[-0.1cm]
%\end{tabular}
\caption{\label{tab:segment.pairs} Comparative performance of classifiers on segment pair data}
\end{table}

%\begin{table}
%\begin{tabular}{ccccccc}
%&Log. Reg. & Lin. SVM & SVM & RF & $k$NN &MLP\\
%\toprule
%&\multicolumn{6}{c}{100  Dimensions, 200  Training samples}\\
%\cmidrule{2-7}
%\cmidrule{2-9} Original Data & 0.534 & 0.532 & 0.470 & 0.336 & 0.355&0.506\\
%\cmidrule{1-7}
%Transformed Data&  0.062 & 0.062 & 0.062 & 0.059 & 0.060&0.062\\
%\midrule
%&\multicolumn{6}{c}{100  Dimensions, 400  Training samples}\\
%\cmidrule{2-7}
%\cmidrule{2-9} Original Data & 0.541 & 0.538 & 0.388 & 0.076 & 0.255&0.499\\
%\cmidrule{1-7}
%Transformed Data&  0.015 & 0.015 & 0.015 & 0.015 & 0.014&0.015\\
%\midrule
%&\multicolumn{6}{c}{100  Dimensions, 1000  Training samples}\\
%\cmidrule{2-7}
%\cmidrule{2-9} Original Data & 0.480 & 0.481 & 0.173 & 0.025 & 0.122&0.009\\
%\cmidrule{1-7}
%Transformed Data&  0.010 & 0.010 & 0.010 & 0.010 & 0.009&0.010\\
%\midrule
%&\multicolumn{6}{c}{100  Dimensions, 2000  Training samples}\\
%\cmidrule{2-7}
%\cmidrule{2-9} Original Data & 0.485 & 0.479 & 0.131 & 0.009 & 0.070&0.0009\\
%\cmidrule{1-7}
%Transformed Data&  0.005 & 0.005 & 0.005 & 0.004 & 0.005& 0.005\\
%\bottomrule
%\\[-0.1cm]
%\end{tabular}
%\caption{\label{tab:segment.pairs.2} Comparative performance of classifiers on segment pair data after transformation by a cumulative sum}
%\end{table}

\subsection{MNIST (Subset)}
To conclude this section we provide (in Table \ref{tab:MNIST}) classification results on a subset of the MNIST digit recognition dataset \citep{lecun1998mnist}, with 10 classes and 100 examples per class for training. To reduce the computation time, we reduced the dimension of the data by transforming the original 28$\times$28 images to 14$\times$14, resulting in 196-dimensional dataset. With this sample size, this reduction had little influence on the performance of classifiers run before diffeomorphic transformation. All classifiers have similar error rates, with a small improvement obtained after diffeomorphic transformation compared to linear classifiers, reaching a final performance comparable to that obtained by random forests and multi-layer perceptrons with two hidden layers. We did observe a glitch in the performance of nonlinear support vector machines, which may result from a poor choice of hyper-parameters (section \ref{sec:classif}) for this dataset.
\begin{table}[h]
\centering
\begin{tabular}{lcccccccc}%
&Log. reg.&lin. SVM&SVM&RF&kNN&MLP1&MLP2&MLP5\\%
\toprule%
\multicolumn{9}{r}{MNIST data, $d=196$, 100 samples per class}\\%
\cmidrule{2-9} Original Data&0.128&0.138&0.228&0.111&0.120&0.122&0.111&0.127\\%
Transformed Data&0.113&0.120&0.325&0.129&0.094&0.110&0.110&0.124\\%
\bottomrule%
\end{tabular}%

%\begin{tabular}{cccccccc}
%&Log. Reg. & Lin. SVM & SVM & RF & $k$NN & MLP\\
%\toprule
%&\multicolumn{6}{c}{100  Dimensions, 1500  Training samples}\\
%\cmidrule{2-7}
%\cmidrule{2-9} Original Data & 0.100 & 0.121 & 0.093 & 0.070 & 0.056 & 0.065\\
%\cmidrule{1-7}
%Transformed Data&  0.050 & 0.051 & 0.050 & 0.046 & 0.044 &0.067\\
%\bottomrule
%\\[-.1cm]
%\end{tabular}
\caption{\label{tab:MNIST} Comparative performance of classifiers on a subset of the MNIST dataset.}
\end{table}

\subsection{Discussion}
The results presented in this section are certainly encouraging, and demonstrate that the proposed algorithm has, at least, some potential.
We however point out that the classifiers that were used in our experiments were run ``off-the-shelves,'' using the setup described in section \ref{sec:classif}. There is no doubt that, after some tuning up specific to each dataset, each of these classifiers could perform better. Such an analysis was not our goal here, however, and  the classification rates that we obtained must be analyzed with this in mind. We just note that we also used the exact same version of  diffeomorphic learning for all datasets, with hyper-parameters values described in section \ref{sec:param}.

We have included kernel-based support vector machines in our comparisons. These classifiers indeed share with the one proposed here the use of a kernel trick to reduce an infinite dimensional problem to one with a number of parameters comparable to the size of the training set. Some notable differences exist however.  The most fundamental one lies in the role that is taken by the kernel in the modeling and computation. For SVMs and other kernel methods, the kernel provides, or is closely related to, the transformation of the data into a feature space. It indeed defines the inner product in that space, and can itself be considered as an infinite dimensional representation of the data, through the standard mapping $x \mapsto \ka(x, \cdot)$. Kernel methods then implement linear methods in this feature space, such as computing a separating hyperplane for SVMs. For the diffeomorphic classification method that is described here, the kernel is used as a computational tool, similar to the way it appears in spline interpolation. The fundamental concept here is the Hilbert space $V$ and its norm, which is, in the case of a $C^k$ Mat\'ern kernel, a Hilbert Sobolev space of order $k+d/2$. In small dimensions, one could actually directly discretize this norm on a finite grid (using finite differences to approximate the derivatives), as done in shape analysis for image registration (see, e.g., \cite{bmty05}). The reproducing kernel of $V$ is therefore a mathematical tool that makes explicit the derivation of the discrete representation described in section \ref{sec:reduction} rather than an essential part of the model, as it is for kernel methods such as SVMs.

Another important distinction, which makes the diffeomorphic method deviate from both kernel methods in machine learning and from spline interpolation is that the kernel (or its associated space $V$) is itself part of a nonlinear dynamical evolution where it intervenes at every time. This dynamical aspect of the approach constitutes a strong deviation from SVMs which applied the kernel transformation once before implementing a linear method. It is, in this regard, closer to feed-forward neural network models in that it allows for very large non-linear transformations of the data to be applied prior to linear classification. Unlike neural networks, however, diffeomorphic learning operates the transformations within the original space containing the data. 

As mentioned in section \ref{sec:classif}, we ran the optimization algorithm until numerical convergence, although it is clear from monitoring classification over time that classes become well separated early on while the estimation of the optimal transformation typically takes more time. If one is not interested in the limit diffeomorphism (which can be of interest for modeling purposes), significant computation time may be saved by stopping the procedure earlier, using, for example a validation set.

\section{Setting Hyper-Parameters}
\label{sec:param}
The diffeomorphic classification method described in this paper requires the determination of several hyper-parameters that we now describe with a discussion of how they were set in  our experiments. 

\begin{enumerate}[label=(\alph*),leftmargin=20pt]
\item Kernel scale parameter. While the kernel can be any function of positive type and therefore considered as an infinite dimensional hyper-parameter, we assume that one uses a kernel such as those described in equations \eqref{eq:gauss.k} and \eqref{eq:matern.k} (our experiments use the latter with $k=3$) and focus on the choice of the scale parameter $\rho$. In our implementation, we have based this selection on two constraints, namely that training data should tend to ``collaborate'' with some of their neighbors from the same class while data from different classes should ``ignore'' each other. This resulted in the following computation.
\begin{enumerate}[label=(\roman*)]
\item To address the first constraint, we evaluate, for each data point, the fifth percentile of its distance to other points from the same class. We then define $\rho_1$ to be the 75th percentile of these values.
\item For the second constraint, we define $\rho_2$ as the 10th percentile of the minimum distance between each data point and its closest neighbor in an other class. 
\end{enumerate}
We finally set $\rho = \rho_0 = \min(\rho_1, \rho_2)$. Table \ref{tab:rhos} provides a comparison of the performances of the algorithm for values of $\rho$ that deviate from this default value, showing that, in most cases, this performance obtained with $\rho_0$ is not too far from the best one. The only exception was found with the Xor dataset, for which a significant improvement was obtained using a large value of $\rho$. Note that in Tables \ref{tab:rhos} and \ref{tab:Ts}, all experiments in the same row were run with the same training and test sets.
\begin{table}[h]
\centering
\begin{tabular}{lcccccccc}
Dataset & Samples/class & $0.25\rho_0$ & $0.5\rho_0$ & $0.75\rho_0$ & $\rho_0$ & $1.5\rho_0$ & $2\rho_0$ & $4\rho_0$\\
\toprule
Sph. Layers& 250 & 0.238 & 0.199&0.188&0.175&0.165&0.183&0.212\\
\midrule
RBF & 100 &0.170 & 0.107 & 0.085 & 0.083 & 0.070& 0.063& 0.063\\ 
\midrule
Tori ($d=10$) & 100 & 0.285 & 0.233 & 0.207 & 0.165 & 0.163 & 0.175 & 0.194\\
\midrule
M. of G. & 100 & 0.322 & 0.221 & 0.156 & 0.135 & 0.109 & 0.103 & 0.124\\
\midrule
Xor & 100 & 0.473 & 0.108 & 0.090 & 0.092 & 0.093 & 0.062 & 0.01\\
\midrule
Seg. Length & 250 & 0.164 & 0.143 & 0.114 & 0.097 & 0.093 & 0.093 &0.86\\
\bottomrule
\end{tabular}
\caption{\label{tab:rhos} Performance comparison when the kernel scale deviates from the default value $\rho_0$, assessed on the spherical layers, RBF, Tori, mixture of Gaussians, Xor and segment length datasets.}
\end{table}

\item Regularization weight. A second important parameter is the regularization weight, $\sigma^2$ in Equation \ref{eq:cost}. In our experiments, this parameter is adjusted on line during the minimization algorithm, starting with a large value, and slowly decreasing it until the training error reaches a target, $\delta$. While this just replaces a parameter by another, this target value is easier to interpret, and we used $\delta = 0.005$ in all our experiments (resulting de facto in a vanishing training error at the end of the procedure in all cases). 

\item $\ell^2$ penalty on logistic regression. This parameter, denoted $\lambda$ in Equation \ref{eq:cost}, was taken equal to 1 in all our experiments.

\item Time discretization. The value of $T$ in equation \eqref{eq:cost.red.2} may also impact the performance of the algorithm, at least for small values, since one expects the model to converge to its continuous limit for large $T$. This expectation is confirmed in Table \ref{tab:Ts}, which shows that the classification error rates obtained with the value chosen in our experiments, $T=10$, was not far from the asymptotic one. In some cases, the error rates obtained for smaller value of $T$ may be slightly better, but the difference is marginal.
\begin{table}[h]
\centering
\begin{tabular}{lcccccccc}
Dataset & Samples/class & $T=1$ & $T=2$ & $T=4$ & $T=10$ & $T=20$ & $T=40$ & $T=100$\\
\toprule
Sph. Layers& 250 & 0.199 & 0.199&0.190&0.197&0.197&0.197&0.197\\
\midrule
RBF & 250 &0.089 & 0.089 & 0.085 & 0.083 & 0.083& 0.083& 0.083\\ 
\midrule
Tori ($d=10$) & 100 & 0.243 & 0.160 & 0.170 & 0.177 & 0.186 & 0.195 & 0.197\\
\midrule
M. of G. & 100 & 0.178 & 0.169 & 0.141 & 0.119 & 0.114 & 0.111 & 0.109\\
\midrule
Xor & 100 & 0.478 & 0.179 & 0.079 & 0.043 & 0.042 & 0.040 & 0.039\\
\midrule
Seg. Length & 250 & 0.140 & 0.117 & 0.100 & 0.094 & 0.092 & 0.091 &0.091\\
\bottomrule
\end{tabular}
\caption{\label{tab:Ts} Performance comparison for various values of the number of discretization steps, $T$,  assessed on the spherical layers, RBF, Tori, mixture of Gaussians, Xor and segment length datasets.}
\end{table}
\end{enumerate}

\section{Conclusion}
\label{sec:discussion}
In this paper, we have introduced the concept of diffeomorphic learning and provided a few illustrations of its performance on simple, but often challenging, classification problems. On this class of problems, the proposed approach appeared quite competitive among other classifiers used as comparison. Some limitations also appeared, regarding, in particular, the scalability of the method, for which we have provided some options that will be explored in the future.

We have only considered, in this paper, applications of diffeomorphic learning to classification problems. Extensions to other contexts will be considered in the future. Some, such as regression, may be relatively straightforward, while others, such as clustering, or dimension reduction should require additional thoughts, as the obtained results will be highly dependent on the amount of metric distortion allowed in the diffeomorphic transformation.

 \bibliographystyle{plain}
%\nocite{*}
\bibliography{diffLearningRevision}

\end{document}